\documentclass[preprints,article,accept,moreauthors]{mdpi}

\usepackage{makecell}
\usepackage[export]{adjustbox}
\usepackage{placeins}

\firstpage{1}
\pubvolume{1}
\issuenum{1}
\articlenumber{0}
\pubyear{2026}
\copyrightyear{2026}
\externaleditor{Pierluigi Siano}
\datereceived{27 July 2026}
\daterevised{4 September 2026}
\dateaccepted{8 September 2026}
\datepublished{ }

\pdfoutput=1

\Title{CoMa: Contextual Massing Generation with \linebreak Vision-Language Models}

\Author{Evgenii Maslov $^{1,}$*, Alexandra Vabnits $^{2}$, Vladimir Vorona $^{3}$, Anastasia Antsiferova $^{1,}$*, Valentin Khrulkov $^{1,2}$, \linebreak Anastasia Volkova $^{1}$, Anton Gusarov $^{1}$, Andrey Kuznetsov $^{1,4}$ and Ivan Oseledets $^{5}$}

\AuthorNames{Evgenii Maslov, Alexandra Vabnits, Vladimir Vorona, Anastasia Antsiferova, Valentin Khrulkov,  Anastasia Volkova, Anton Gusarov, Andrey Kuznetsov and Ivan Oseledets}

\address{
$^{1}$ \quad FusionBrain Lab, 
Artificial Intelligence Research Institute, 123112 Moscow, Russia; khrulkov.v@gmail.com (V.K.); a.volkova@fusionbrainlab.com (A.V.); a.gusarov@fusionbrainlab.com (A.G.); \linebreak kuznetsov@fusionbrainlab.com (A.K.) 
\\
$^{2}$ \quad Research Center of the Artificial Intelligence Institute, Innopolis University, 420500 Innopolis, Russia; al.vabnits@innopolis.ru  \\
$^{3}$ \quad ITMO University, Institute of Design and Urban Studies, 197101 Saint Petersburg, Russia; v.vorona@itmo.ru \\
$^{4}$ \quad Multimodal Industrial GenAI Lab, Innopolis University, 420500 Innopolis, Russia\\
$^{5}$ \quad Artificial Intelligence Research Institute, 123112 Moscow, Russia; ivan.oseledets@gmail.com}

\corres{Correspondence: em.maslov@fusionbrainlab.com (E.M.); av.antsiferova@fusionbrainlab.com (A.A.)
}

\addhighlights{yes}
\renewcommand{\addhighlights}{%

\textbf{What are the main findings?}
\begin{itemize}[labelsep=2.5mm,topsep=-3pt]
\item Multimodal training (vector + map + 3D-view context) \textcolor{black}{consistently} improves contextual relevance of generated massings compared to any single modality. 
\item Larger VLM capacity (2B to 8B parameters) consistently improves contextual relevance metrics.
\end{itemize}\vspace{3pt}

\textbf{What are the implications of the main findings?}
\begin{itemize}[labelsep=2.5mm,topsep=-3pt]
\item Multimodal approaches from this paper enable architects to receive site-aware designs without manual context specification.
\item Mid-scale models offer the most practical trade-off for \textcolor{black}{architectural workflows} due to the marginal gains relative to computational cost.
\end{itemize}
}

\abstract{Context-aware building massing is an important early-stage design task: given a site for buildings, a generated massing should not only fit the target parcel, but also relate to the scale, density, and morphology of its surrounding urban fabric. This task is naturally multimodal, since the target output should remain structured and editable, while the surrounding context, including other buildings or roads, can be represented as vector geometry, map imagery, or three-dimensional views. 
In this paper, we study contextual massing generation using vision-language models (VLMs) and analyze their performance on this task across different context modalities during training and inference.
We assemble an experimental dataset of 12,845 Melbourne massings with parcel contours, structured 3D geometry, neighboring buildings, top-down views, and multi-view 3D context images. \textcolor{black}{We also introduce a learned contextual relevance metric for evaluating whether generated massings are morphologically compatible with their surrounding context.} Using Qwen3-VL models, we compare no-context, unimodal-context, and multimodal-context training regimes and evaluate inference performance under controlled combinations of modalities and amounts of context. The results show that model size strongly affects generation quality, multimodal training improves the use of individual modalities, and multimodal inference provides a stronger contextual signal than isolated context inputs.
}


\begin{document}

\section{Introduction}

Early-stage urban planning and architectural design require rapid exploration of development alternatives before detailed facade, structural, or~interior decisions are available. At~this stage, planners and designers consider various requirements for the project: building density, floor area, height, site occupation, functional program, and~compatibility with the surrounding urban fabric. Building massing is the key representation for this phase: it abstracts a proposal into editable three-dimensional volumes, footprints, and~height relationships that can be inspected, compared, and~further analyzed in GIS, simulation, and~digital-twin workflows
~\cite{rhee2020integrating,wang2020algorithmic}.

Contextual massing generation addresses a more specific and practically important question: how to generate the three-dimensional form of a new building so that it responds to the surrounding urban fabric. In~real design practice, a~proposal is expected not only to satisfy parcel-level geometric constraints, but~also to maintain a meaningful relationship with nearby buildings in terms of scale, height, volume, and~morphological character. This requirement is central to the creation of coherent urban environments, yet it is difficult to formalize in a computationally tractable way. Unlike reconstruction accuracy or image similarity, morphological contextual compatibility has no single established definition or evaluation protocol. In~this work, we take a step toward addressing this gap by introducing metrics that estimate how well a generated massing corresponds to the surrounding built~context.

Computational approaches to massing and urban form generation have traditionally relied on procedural modeling, parametric design, and~optimization-based workflows, which provide explicit control, but~often require expert-authored rules, predefined design variables, or~formalized objective functions~\cite{parish2001procedural,wang2020algorithmic,perezmartinez2023methodology}. Recent learning-based methods expand this design space through graph and vector representations for structured layouts~\cite{xu2021blockplanner,he2023globalmapper}, architectural generation conditioned on sketches or images~\cite{tono2024vitruvio}, and~multimodal or diffusion-based urban synthesis~\cite{zhuang2024texttocity,he2025stepwise}. However, these works typically address related but narrower tasks, such as two-dimensional layout generation, sketch-based reconstruction, image synthesis, or~scene-level urban generation. They do not provide a general approach to producing editable parcel-level three-dimensional massing, nor do they directly evaluate whether the generated building forms are consistent with the surrounding urban~context.

Vision-language models are not strictly required for massing generation: conventional procedural, parametric, and~optimization-based methods remain effective when the design space, constraints, and~evaluation criteria can be explicitly formalized. However, these methods typically rely on predefined parameterizations, expert-authored rules, or~formal objective functions, making it difficult to incorporate contextual information that is heterogeneous or not readily reducible to numerical criteria. VLMs offer complementary capabilities: they can jointly process natural-language instructions, reference images, and~structured descriptions of the surrounding environment while generating editable geometry encoded as a structured vector language~\cite{llm_review}. This makes them a promising candidate for investigating context-aware massing generation, particularly when design intent and urban context are provided through multiple modalities. Our aim is therefore not to establish that VLMs are necessary or universally superior, but~to evaluate their potential for this setting and determine how effectively they use different representations of urban~context.

At the same time, recent spatial-reasoning benchmarks show that multimodal models still struggle with quantitative spatial relations, coordinate fidelity, and~three-dimensional spatial understanding~\cite{zhan2025open3dvqa,xu2025spatialbench}. This raises two open questions: whether combining context modalities during training helps VLMs better use each individual modality, and~whether combining modalities during inference improves generation quality compared with isolated context inputs. More broadly, we ask how effectively VLMs can use urban context for structured 3D massing generation, and~how their performance depends on the modality and amount of contextual information available during training and~inference.

This paper presents two key contributions:
(1) a contextual relevance metric for estimating whether
generated building massing corresponds to the surrounding
urban fabric in terms of geometric and morphological relationships;
and (2) a systematic study of VLMs for contextual massing generation, analyzing whether multimodal training improves the use of individual context modalities and whether multimodal inference provides gains over isolated context~inputs. The experiment code, as well as the dataset, model weights, and inference results are publicly available and can be accessed via links provided in Supplementary Materials.

Our work shifts the focus from pursuing a single best-performing generator to understanding how vision-language models use multimodal urban context in architectural massing tasks, highlighting both their potential and their limitations for context-aware design~generation.

\section{Related~Works}
\unskip

\subsection{Procedural, Parametric, and~Morphology-Aware~Massing}

Procedural and parametric methods are a natural starting point for urban massing because they encode geometry through explicit rules and controllable parameters. Classical procedural city modeling established grammar-driven pipelines in which roads, parcels, and~buildings are generated from global and local rules~\cite{parish2001procedural}. Interactive procedural street generation extended this paradigm by allowing users to guide street networks through editable tensor fields and graph-level operations~\cite{chen2008interactive}. These systems remain influential because they provide interpretable control over urban form, but~their realism and adaptability depend heavily on the quality of manually designed~rules.

At the building scale, parametric and evolutionary approaches have been used to generate architectural massing models through additive and subtractive form operations~\cite{wang2020algorithmic}. Related performance-based methods combine massing and facade design in evolutionary optimization loops, allowing form alternatives to be evaluated against environmental or design objectives~\cite{zhang2021synergy}. In~urban planning, generative-design workflows formulate subdivision, typology allocation, and~planning-indicator balancing as computational search problems~\cite{perezmartinez2023methodology}. These approaches are strong when constraints and objectives can be clearly specified, but~they are less suited to settings where requirements include implicit contextual preferences, visual compatibility, or~partially formalized design intentions. A~related line in urban morphology emphasizes higher-level control through types, fabrics, and~morphotopes rather than only low-level scalar parameters, suggesting an important direction for planning-oriented controllability even when the generative mechanism itself is not based on neural models~\cite{lipovka2021morphological}.

\subsection{Graph, Vector, and~Layout-Based Generative~Models}

Learning-based layout generation has increasingly moved from raster outputs toward graph and vector representations. This shift is important because architectural and urban layouts require topology, adjacency, containment, and~boundary relationships to remain explicit. House-GAN introduced a graph-constrained generative adversarial model for residential layout generation, showing how room relations can be encoded directly in the generative process~\cite{nauata2020housegan}. House-GAN++ later refined this line through adversarial layout refinement networks~\cite{nauata2021houseganpp}. Graph2Plan similarly converts layout graphs into floor-plan configurations, demonstrating the value of relational structure for architectural generation~\cite{hu2020graph2plan}.

At the urban scale, BlockPlanner proposed a vectorized graph representation for city-block generation, with~a two-level structure designed to preserve both block topology and land-lot geometry~\cite{xu2021blockplanner}. GlobalMapper addressed arbitrary-shaped urban layout generation, focusing on the challenge of producing plausible layouts under irregular block boundaries~\cite{he2023globalmapper}. These methods are closely related to contextual massing because they emphasize structured geometry rather than only visual appearance. However, their main focus is two-dimensional layout generation, while contextual massing generation requires editable three-dimensional geometry whose scale and form respond to the surrounding built~environment.

Building-scale three-dimensional generation has also been studied under sketch- and image-conditioned formulations. Vitruvio uses a conditional variational autoencoder to generate building meshes from a single perspective sketch~\cite{tono2024vitruvio}. This is relevant because it treats building geometry as a learnable output rather than a purely rule-based object. However, sketch-driven mesh generation differs from contextual massing generation, where the input is a target site and surrounding urban context rather than a manually drawn sketch. Work on integrating building footprint prediction with downstream massing further highlights the importance of urban context, but~it does not examine how surrounding built form can be represented and used to guide morphologically compatible three-dimensional massing generation~\cite{rhee2020integrating}.

\subsection{Diffusion and Multimodal Urban~Generation}

Diffusion models have recently expanded the range of generative methods available for urban and architectural design. Text-to-City applies latent diffusion to controllable three-dimensional urban block generation, showing that textual prompts can guide the synthesis of urban-scale form~\cite{zhuang2024texttocity}. Diffusion has also been used for urban-planning image synthesis, where satellite-like imagery is generated under land-use and infrastructure conditions~\cite{wang2025urbanplanningdiffusion}. These approaches demonstrate the expressive power of diffusion models, but~they often prioritize raster or rendered outputs rather than structured massing geometry that can be directly measured and~edited.

More workflow-oriented multimodal systems attempt to bridge generative AI and professional urban-design processes. A~stepwise multimodal diffusion framework decomposes urban design into road and land-use planning, building layout planning, and~rendering, which better reflects iterative planning practice than single-step generation~\cite{he2025stepwise}. Urban Architect introduces steerable three-dimensional urban scene generation with layout priors, combining text-driven control with spatial structure~\cite{lu2024urbanarchitect}. These systems are important because they show that multimodal generation can support urban scenario creation. Nevertheless, they are not designed to analyze how generative models use surrounding morphological context when producing editable parcel-level massing~geometry.

\subsection{Vision-Language Models for Structured Spatial~Generation}

Vision-language models provide a different capability profile from diffusion models. Instead of generating only images or implicit scene representations, they can process images and text while producing structured textual outputs. This is relevant for massing generation because geometric data can be serialized into text-based formats such as JSON, SVG-like code, or~coordinate sequences. Pix2Seq demonstrated that object detection can be framed as sequence generation over coordinates and classes~\cite{chen2021pix2seq}. Perceiver IO generalized the idea of multimodal structured prediction through an architecture designed for flexible input and output structures~\cite{jaegle2021perceiverio}. StarVector showed that multimodal models can generate scalable vector graphics code from images and text, suggesting that vector geometry can be treated as a language-like output~\cite{rodriguez2023starvector}.

Recent VLMs further motivate this direction because they combine visual perception, instruction following, and~structured response generation~\cite{bai2025qwen25vl}. These properties make VLMs a plausible baseline for contextual massing generation, where visual urban context, textual requirements, and~coordinate-based geometry must be handled together. However, massing generation stresses capabilities that remain weak in current models. Open3DVQA shows that multimodal large language models still face difficulties in open-space three-dimensional spatial reasoning~\cite{zhan2025open3dvqa}. SpatialBench similarly reports limitations in spatial cognition across multimodal large language models~\cite{xu2025spatialbench}. EarthSpatialBench extends this concern to Earth imagery, indicating that geospatial reasoning remains a challenging domain for current multimodal systems~\cite{xu2026earthspatialbench}. These findings motivate a systematic study of VLMs in contextual massing generation, focusing not only on output validity, but~also on whether models can use different representations of surrounding urban context to produce morphologically compatible 3D~forms.

\section{Task~Definition}
\label{sec:task}

We study context-conditioned massing generation: given a development site and its surrounding urban environment, a~generative model must produce the geometry of the building or group of buildings to be erected on that site. Formally, we consider a generator
\begin{equation}
\label{eq:generation_task}
g: S \times C \rightarrow M,
\end{equation}
where $S$ is the space of development-site contours, $C$ is the space of contextual information about the surrounding built environment, and~$M$ is the space of output massing~geometries.

\begin{figure}[H]

\includegraphics[width=\textwidth]{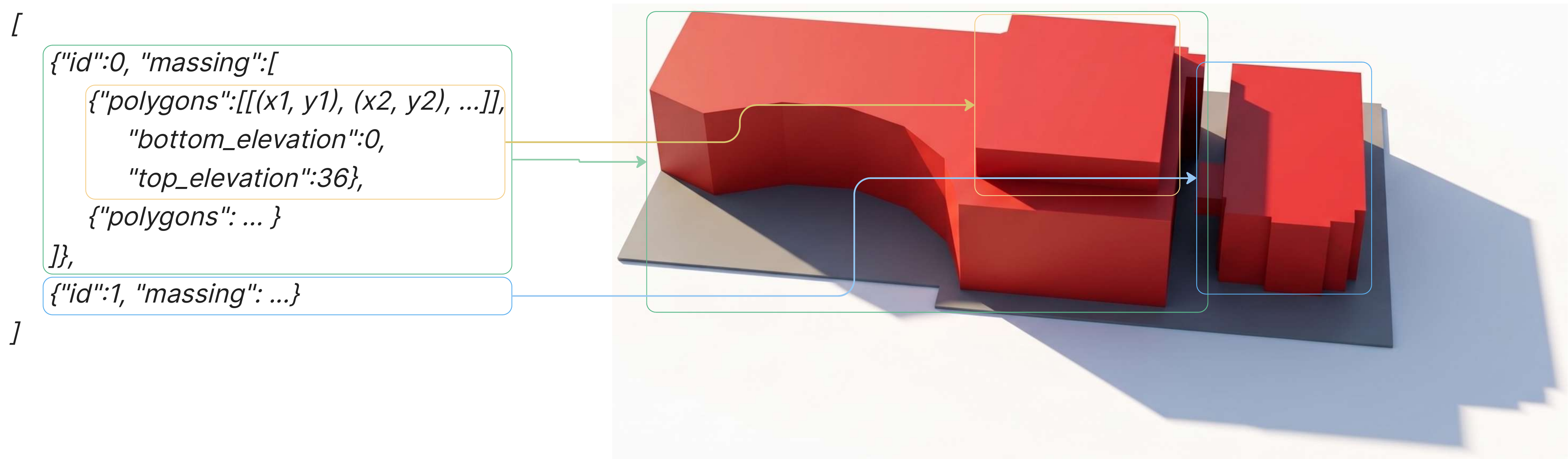}
\caption{Our
 implementation of the LoD 1.3 abstraction for a machine-readable representation of a massing as a JSON~structure.}
\label{fig:json_vector_representation}
\end{figure}

Following the classification of level of detail in~\cite{biljecki2016improved}, the~output massing $m \in M$ corresponds to the LoD~1.3 abstraction for three-dimensional building models: buildings are represented as block models with vertically differentiated roof and body parts but without detailed facade semantics. We encode this abstraction as a hierarchical object
\begin{equation}
\label{eq:massing}
m=(b_1,\dots,b_K),\quad
b_k=((e_{k,1},\dots,e_{k,L_k})),\quad
e_{k,l}=(z^{\downarrow}_{k,l},z^{\uparrow}_{k,l},(P_{k,l,1},\dots)),
\end{equation}
where each building $b_k$ is a stack of vertical extrusions $e_{k,l}$ defined by a bottom elevation $z^{\downarrow}$, a~top elevation $z^{\uparrow}$, and~a set of footprint polygons $P$. Each polygon is an ordered list of planar points. Formally, each footprint polygon is represented as
\begin{equation}
\label{eq:polygon}
P_{k,l,r}=(p_{k,l,r,1},\dots,p_{k,l,r,N_{k,l,r}}),\qquad
p_{k,l,r,i}=(x_{k,l,r,i},y_{k,l,r,i})\in\mathbb{R}^2,
\end{equation}
where \(r\) indexes polygons within an extrusion and \(N_{k,l,r}\) is the number of points in polygon \(P_{k,l,r}\). In~practice, the~hierarchical structure is implemented as a JSON object, schematically shown in Figure~\ref{fig:json_vector_representation}.

The site contour $s \in S$ is represented as a set of polygons,
\begin{equation}
\label{eq:site_contour}
s=(P_1,\dots,P_N),
\end{equation}
where each polygon $P_i$ has the same point-based format as the footprint polygons used inside massing extrusions. In~the implemented JSON representation, a~point is a tuple of two floating-point values, a~polygon is a list of such tuples, and~the site contour is a list of~polygons.

For both the site contour and massing geometry, polygon coordinates are expressed in a relative local coordinate system. The~origin of this system is the first point of the first polygon of the site contour. All polygon coordinates are rounded to two decimal places. Since terrain is not considered in this formulation, the~lowest bottom elevation of the output massing is always set to~zero.

The context $c \in C$ contains information about the surrounding built environment. We define it as a collection of context information objects of possibly different types:
\begin{equation}
\label{eq:context}
c=\{c^{(t)}_j \mid t \in \mathcal T,\; j=1,\dots,n_t\},
\end{equation}
where $\mathcal T$ is the set of available context types and $n_t$ is the number of context objects of type~$t$.

Our research question is how the type and amount of surrounding context provided to a VLM affect the quality of generated massings in terms of compatibility with the surrounding urban context. We use this formulation to evaluate how different context-presentation strategies during training and inference affect VLMs' ability to solve contextual massing generation. The~concrete context types, or~modalities, together with the VLM training and inference details, are described later in Section~\ref{sec:experiments}. However, to~conduct this comparison, we need a contextual-relevance metric that quantifies whether a generated massing fits its surroundings. The~design and validation of this metric are introduced~next.

\section{Metrics~Design}
\label{sec:metrics}

A contextual-relevance metric must quantify whether a generated massing fits its surroundings. We build one on an idea from urban-morphology work: a building fits its context when its geometric attributes (height, density, orientation, setbacks, footprint shape) stay close to those of the neighboring buildings, and~an overall fit score can be assembled from such per-attribute comparisons~\cite{biljecki2022morphology,labetski2023metrics}. This section makes that idea operational. We first state what ``fit'' means and how we will choose a metric (Section~\ref{subsec:idea}); we then define a space of candidate metrics built from massing characteristics, non-learnable combinations of them, and~a learnable ensemble (Section~\ref{subsec:space}); and we finally validate, compare, and~interpret these candidates to select the metric used in the rest of the paper (Section~\ref{subsec:analysis}). Throughout, we follow one principle: every modeling choice is justified from data rather than left to the reader to~infer.

\subsection{Contextual Relevance~Idea}
\label{subsec:idea}

We adopt an operational definition of morphological fit: a building fits its context if it behaves as real buildings do there, that is, if~its geometry is statistically indistinguishable from that of a true building placed in the same surroundings. Because~the true functional dependence of geometry on context is unknown, we treat a contextual-relevance metric as a binary classifier that decides whether a massing--context pair is genuine. Choosing a metric then becomes a concrete sub-problem: among many candidate classifiers, find the one that best tells genuine massing--context pairs apart from mismatched ones. Such a classifier can also be inverted---by perturbing the context and checking whether its decision changes---to measure how sensitive the metric is to context, and~hence how much genuine massing--context dependence it~encodes.

We turn this into a measurable selection criterion using a balanced classification benchmark, built from the dataset in Section~\ref{subsec:dataset} via negative sampling. For~every genuine massing--context pair we create a matched negative by replacing its context with a uniformly resampled foreign context, giving equal numbers of positive and negative examples (\mbox{$N$ = 25,690}; 12,845 of each class). These examples are drawn from a shared pool of 12,959 unique context massings: 12,845 have target-site labels and therefore yield positive benchmark pairs, while the remaining $114$ are context-only massings that can appear as surrounding buildings but are not used as targets. A~fixed score needs no fitting, so we rank it directly on the whole benchmark; a learned metric is instead trained and scored under GroupKFold cross-validation grouped by target massing ID, which keeps the positive and negative examples of a site in the same fold and so prevents leakage~\cite{scikit-learn}. We compare candidates by ROC--AUC as the primary measure and the best-threshold $F_1$ score as a secondary, decision-oriented one~\cite{fawcett2006roc,powers2011evaluation}. The~metric that best separates genuine from foreign context on real data is the one we use to later rank trained generators across context~modalities.

\subsection{Metric Space~Definition}
\label{subsec:space}

A metric assigns a fit score to a massing--context pair. Rather than comparing raw geometry, we describe each massing in terms of interpretable characteristics and build metrics on top of them. We then search a space of candidate metrics organized into three increasingly expressive families: (i)~scores built from a single characteristic, (ii)~non-learnable metrics that compare whole feature distributions, and~(iii)~learnable ensembles that combine several scores. We define the characteristics first, then the non-learnable metrics, then the~ensembles.

\subsubsection{Notation}
\label{subsubsec:notation}

We reuse the massing representation of Equation~(\ref{eq:massing}) throughout. A~massing $m$ has $K$ buildings $b_1,\dots,b_K$; building $b_k$ is a stack of extrusions indexed by $l$, and~each extrusion carries footprint polygons indexed by $i$. We write $F(m)=\bigcup_{k,l,i}P_{k,l,i}$ for the horizontal union of all footprint polygons of $m$, and~$F_k=\bigcup_{l,i}P_{k,l,i}$ for the corresponding projected footprint of building $b_k$. The~ground footprint $F_k^{\downarrow}$ is the union of the footprint polygons in the lowest-bottom extrusions of $b_k$; $F^{\downarrow}(m)$ and $F^{\uparrow}(m)$ are the unions, over~buildings, of~their lowest-bottom and highest-top extrusion footprints. For~a building, $z_k^{\downarrow}=\min_l z_{k,l}^{\downarrow}$ and $z_k^{\uparrow}=\max_l z_{k,l}^{\uparrow}$ are its bottom and top elevations; for an extrusion $e$, $z_e^{\downarrow},z_e^{\uparrow}$ and $F_e$ are its bottom elevation, top elevation, and~footprint. We use $A(\cdot)$ for area, $\ell(\cdot)$ for perimeter, $\operatorname{conv}(\cdot)$ for the convex hull, and~$\operatorname{dist}(\cdot,\cdot)$ for the minimum distance between two geometries; $S_m$ is the site polygon associated with $m$ and $\partial S_m$ its boundary. A~characteristic is a scalar function $\phi(\cdot)$ of a massing, and~a score is a metric value assigned to a target--context~pair.

\subsubsection{Characteristics}
\label{subsubsec:characteristics}

We summarize each massing with twelve scalar characteristics $\phi(\cdot)$, organized into three families that mirror how an architect reads a site: how big and dense the buildings are (scale \& density), what their footprints look like (shape), and~how they are turned relative to the street grid (orientation). We choose these three because they are the axes along which urban fabric is conventionally described~\cite{biljecki2022morphology,labetski2023metrics} and because, as~Section~\ref{subsec:analysis} shows, they behave very differently as contextual cues; we reuse the same grouping in every later figure. Variety of geometric detail (variation in heights, elevations, and~corner angles) is itself an aspect of form, so we place these diversity descriptors in the shape family rather than in a separate~group.

For the height and floor-area characteristics, we slice each building into floors of fixed height $h_0=3\,\text{m}$. For~building $b_k$, let $n_{\text{fl},k}=\left\lceil(z_k^{\uparrow}-z_k^{\downarrow})/h_0\right\rceil$ and define floor base levels $h_{k,j}=z_k^{\downarrow}+(j-1)h_0$, $j=1,\dots,n_{\text{fl},k}$. The~scale \& density characteristics---how tall, how intensively built, and~how the footprint sits on the parcel---are
\vspace{6pt}
\begin{align}
\phi_{\text{hgt}}(m) &= \frac{1}{K}\sum_k\left\lceil\frac{z_k^{\uparrow}-z_k^{\downarrow}}{h_0}\right\rceil,
&&\text{mean floor count,} \\
\phi_{\text{far}}(m) &= \frac{1}{A(S_m)}\sum_{k}\sum_{j=1}^{n_{\text{fl},k}} A\!\left(\bigcup_{e\in b_k:\,h_{k,j}\in[z_e^{\downarrow},z_e^{\uparrow})} F_e\right),
&&\text{floor-area ratio,} \\
\phi_{\text{cov}}(m) &= A(F(m))/A(S_m),
&&\text{projected coverage,} \\
\phi_{\text{set}}(m) &= \frac{1}{K}\sum_k \operatorname{dist}\left(F_k^{\downarrow},\partial S_m\right),
&&\text{mean setback.}
\end{align}

In
 $\phi_{\text{far}}$ the inner union is the part of building $b_k$ that intersects floor $j$: it collects the footprints $F_e$ of all extrusions $e$ whose elevation span $[z_e^{\downarrow},z_e^{\uparrow})$ contains the floor level $h_{k,j}$, so the double sum is the total floor area stacked over the site. The~coverage characteristic uses the horizontal union of all extrusion footprints, matching the implemented projected footprint; it is therefore a projected coverage measure rather than a strictly ground-level coverage when upper extrusions overhang or differ from the lowest footprint. Intuitively, this family answers ``is the proposal as tall, as~dense, and~as set back from the street as its neighbours?''.

The shape characteristics describe footprint form and the variety of geometric detail. The~form descriptors are
\begin{align}
\phi_{\text{elo}}(m) &= 1-\lambda_{\min}/\lambda_{\max},
&&\text{elongation,} \\
\phi_{\text{circ}}(m) &= \frac{1}{K}\sum_k 4\pi A(F_k)/\ell(F_k)^2,
&&\text{mean circularity,} \\
\phi_{\text{conc}}(m) &= 1-A(F(m))/A\!\left(\operatorname{conv} F(m)\right),
&&\text{concavity ratio,} \\
\phi_{\text{stp}}(m) &= A(F^{\uparrow}(m))/A(F^{\downarrow}(m)),
&&\text{stepback ratio,}
\end{align}
where in $\phi_{\text{elo}}$, $\lambda_{\max}\ge\lambda_{\min}$ are the eigenvalues of the covariance matrix of the exterior vertices of $F(m)$. We define the footprint convexity (solidity) $\operatorname{cvx}(F)=A(F)/A(\operatorname{conv}F)$, so the concavity ratio is its whole-massing complement, $\phi_{\text{conc}}=1-\operatorname{cvx}(F(m))$; the per-building convexity is reused by the Fr\'echet metric below. The~implemented stepback ratio clips $A(F^{\uparrow}(m))/A(F^{\downarrow}(m))$ to $[0,1]$. The~variety of detail is measured by three diversity descriptors, each a coefficient of variation $\operatorname{cv}(X)=\operatorname{std}(X)/\operatorname{mean}(X)$ of a set of per-element values $X$:
\begin{align}
\phi_{\text{exd}}(m) &= \operatorname{cv}\!\left(\{\,z_e^{\uparrow}-z_e^{\downarrow}\,\}_{e}\right),
&&\text{extrusion-height diversity,} \\
\phi_{\text{eld}}(m) &= \operatorname{cv}\!\left(\{\,z_e^{\uparrow}\,\}_{e}\right),
&&\text{elevation diversity,} \\
\phi_{\text{agd}}(m) &= \operatorname{cv}\!\left(\{\,\measuredangle_p\,\}_{p}\right),
&&\text{corner-angle diversity,}
\end{align}
where $e$ ranges over all extrusions of $m$ and $\measuredangle_p$ over all interior corner angles of all footprint polygons; in the implementation, corner angles are computed in degrees and rounded to the nearest integer before the coefficient of variation is taken. These three diversities are the informative survivors of a broader set of generic complexity descriptors---element counts, side-length and shape ratios, and~the same diversities under other reductions---measures we did not design specifically for contextual relevance but tried in case they helped; only these three separated genuine from foreign context on the benchmark, so they are the only ones we keep as characteristics. Intuitively, the shape family answers ``are the footprints as round/elongated, as~perforated, and~as internally varied as the neighbours'?''.

The single orientation characteristic $\phi_{\text{dir}}$ averages, over~the $K$ buildings, the~direction $\theta(F_k)\in[0,\pi)$ of the longest edge of the minimum rotated rectangle bounding footprint $F_k$:
\begin{equation}
\label{eq:dir}
\phi_{\text{dir}}(m)=\frac{1}{K}\sum_k \theta(F_k)\in[0,\pi),
\end{equation}
where the half-open range $[0,\pi)$ reflects the $180^{\circ}$ rotational symmetry of a footprint. This scalar direction characteristic follows the implementation directly: per-building angles are computed in degrees, folded into the half-circle, and~then averaged arithmetically; the formula above gives the equivalent radian notation. The~distribution metrics below use a doubled-angle representation for circular comparisons, but~the scalar $\mathrm{Dir}_{\mathrm{near}}$ and $\mathrm{Dir}_{\mathrm{mean}}$ scores use ordinary differences in this folded scalar. Orientation answers ``is the building turned to the same street grid as its neighbours?'' and, as~we show later, is the single most informative~cue.

\subsubsection{Non-Learnable~Metrics}
\label{subsubsec:non_learnable_metrics}

We form non-learnable metrics from these characteristics in three ways: a base score on a single characteristic, a~Fr\'echet comparison of a feature vector, and~a distribution metric specialized for~orientation.

\textbf{Single-characteristic base score.}
Let $\phi(\cdot)$ be a characteristic. For~a target massing $m$ with $Q$ context massings $c_1,\dots,c_Q$, we define
\begin{equation}
\label{eq:relevance}
E(m\mid c_{1:Q})=
\exp\!\left(
-\frac{\left|\phi(m)-\rho(\Phi_c)\right|}{\sigma(\Phi_c)}
\right),
\qquad
\Phi_c=\{\phi(c_q)\}_{q=1}^{Q},
\end{equation}
where the context reduction $\rho$ collapses $\Phi_c$ to a single reference value and $\sigma(\Phi_c)$ is its standard deviation. The~implementation drops context massings whose characteristic is non-finite and requires at least one valid context value; scores with non-finite results are excluded from the benchmark score matrix. We use
\begingroup
\makeatletter\def\f@size{9}\check@mathfonts
\def\maketag@@@#1{\hbox{\m@th\normalsize\normalfont#1}}
\begin{equation}
\rho_{\textsc{mean}}(\Phi_c)=\frac{1}{Q}\sum_{q=1}^{Q}\phi(c_q),
\qquad
\rho_{\textsc{near}}(\Phi_c;\phi(m))=\phi(c_{q^\star}),\quad
q^\star=\arg\min_{q}\left|\phi(c_q)-\phi(m)\right|.
\end{equation}
\endgroup

The
 \textsc{mean} reduction compares the target against the typical context value, whereas the \textsc{nearest} reduction matches it to the closest single context value. \textsc{nearest} is essential when a characteristic is multi-peaked over the context, that is, when its distribution across the context buildings has several distinct peaks rather than one. The~clearest example is street-grid orientation: neighboring blocks align to one of two roughly perpendicular grid directions, so the mean angle falls in the empty gap between the two peaks and is uninformative, while the nearest-peak value is exactly the cue a fitting building should match. Here, multi-peaked refers to modes of the value distribution, not the input modalities of Section~\ref{sec:task}.

We do not pair every characteristic with both reductions; for each characteristic we keep the reduction that scores higher in a preliminary single-characteristic sweep on the benchmark (Section~\ref{subsec:analysis}). The~pattern is systematic: characteristics that are multi-peaked over the context or otherwise best matched to a single neighbor---orientation, circularity, and~the three diversities---use $\rho_{\textsc{near}}$, whereas the single-peaked scale, density, and~form characteristics---FAR, coverage, height, elongation, concavity, stepback, and~setback---are best compared with the typical context value and use $\rho_{\textsc{mean}}$. We name each resulting score by its characteristic with the reduction as a subscript, and~we reuse these symbols verbatim in every figure so that a plot label identifies its formula. For~example, the~orientation score with the nearest reduction is
\[
\mathrm{Dir}_{\mathrm{near}}=E(m\mid c_{1:Q})\ \text{with}\ \phi=\phi_{\text{dir}},\ \rho=\rho_{\textsc{near}}.
\]

\textls[-15]{The nine \emph{core} single-characteristic scores are $\mathrm{Dir}_{\mathrm{near}},\mathrm{FAR}_{\mathrm{mean}},\mathrm{Cov}_{\mathrm{mean}},\mathrm{Hgt}_{\mathrm{mean}},\mathrm{Circ}_{\mathrm{near}}$, $\mathrm{Elo}_{\mathrm{mean}}$, $\mathrm{Conc}_{\mathrm{mean}},\mathrm{ExD}_{\mathrm{near}},$ and $\mathrm{ElD}_{\mathrm{near}}$---the best reduction in each of the nine strongest characteristics, one per characteristic and spanning all three families. Two characteristics, direction and height, also carry independent signal under their other reduction, so we additionally retain $\mathrm{Dir}_{\mathrm{mean}}$ and $\mathrm{Hgt}_{\mathrm{near}}$; with the three weaker scalars $\mathrm{Set}_{\mathrm{mean}},\mathrm{Stp}_{\mathrm{mean}},$ and $\mathrm{AgD}_{\mathrm{near}}$ these make up a pool of extra scalar scores used only by the largest ensemble (Table~\ref{tab:metric-inventory}).}

\begin{table}[H]
\centering
\caption{The $17$ benchmark scores grouped into the three nested feature sets (read cumulatively: \textsc{Core}\,$\subset$\,\textsc{Core+Dist}\,$\subset$\,\textsc{All}). Single-characteristic scores are Equation~(\ref{eq:relevance}) with the listed characteristic and reduction; the symbols are reused verbatim in every~figure.}
\label{tab:metric-inventory}
\begin{tabularx}{\textwidth}{L l l}
\toprule
\textbf{Symbol} & \textbf{Characteristic/Definition} & \textbf{Family} \\
\midrule
\multicolumn{3}{l}{\textsc{Core}: nine single-characteristic scalars}\\
$\mathrm{Dir}_{\mathrm{near}}$  & $\phi_{\text{dir}}$, nearest & orientation \\
$\mathrm{FAR}_{\mathrm{mean}}$  & $\phi_{\text{far}}$, mean   & scale \& density \\
$\mathrm{Cov}_{\mathrm{mean}}$  & $\phi_{\text{cov}}$, mean   & scale \& density \\
$\mathrm{Hgt}_{\mathrm{mean}}$  & $\phi_{\text{hgt}}$, mean   & scale \& density \\
$\mathrm{Circ}_{\mathrm{near}}$ & $\phi_{\text{circ}}$, nearest & shape \\
$\mathrm{Elo}_{\mathrm{mean}}$  & $\phi_{\text{elo}}$, mean   & shape \\
$\mathrm{Conc}_{\mathrm{mean}}$  & $\phi_{\text{conc}}$, mean   & shape \\
$\mathrm{ExD}_{\mathrm{near}}$  & $\phi_{\text{exd}}$, nearest & shape \\
$\mathrm{ElD}_{\mathrm{near}}$  & $\phi_{\text{eld}}$, nearest & shape \\
\midrule
\multicolumn{3}{l}{$+$ distribution scores $\;\rightarrow\;$ \textsc{Core+Dist} ($12$)}\\
$\mathrm{Fr}_{\mathrm{orient}}$  & Equation~(\ref{eq:frechet}), orientation pair & distribution \\
$\mathrm{KDE}_{\mathrm{orient}}$ & Equation~(\ref{eq:kde})            & distribution \\
$\mathrm{Fr}_{\mathrm{multi}}$   & Equation~(\ref{eq:frechet}), full $\mathbf f_b$ & distribution \\
\midrule
\multicolumn{3}{l}{$+$ extra scalars $\;\rightarrow\;$ \textsc{All} ($17$)}\\
$\mathrm{Dir}_{\mathrm{mean}}$  & $\phi_{\text{dir}}$, mean   & orientation \\
$\mathrm{Hgt}_{\mathrm{near}}$  & $\phi_{\text{hgt}}$, nearest & scale \& density \\
$\mathrm{Set}_{\mathrm{mean}}$  & $\phi_{\text{set}}$, mean   & scale \& density \\
$\mathrm{Stp}_{\mathrm{mean}}$  & $\phi_{\text{stp}}$, mean   & shape \\
$\mathrm{AgD}_{\mathrm{near}}$  & $\phi_{\text{agd}}$, nearest & shape \\
\bottomrule
\end{tabularx}
\end{table}

\textbf{Fr\'echet distribution metric.}
A single reduced value cannot tell whether the target reproduces the spread and shape of the context's feature distribution---a context that mixes a few towers with many low blocks, or~one with two competing grid orientations, looks the same to the scalar scheme as a uniform one. The~Fr\'echet metric instead asks whether the target's buildings, taken as a population, are distributed as in the context. For~each building $b$, we form a shape vector that samples a few features from each characteristic family,
\begin{equation}
\mathbf f_b=\big[\underbrace{\log(A+1),\;z^{\uparrow}-z^{\downarrow}}_{\text{scale}},\;\underbrace{\operatorname{cvx},\;\mathrm{circ},\;\mathrm{elo}}_{\text{shape}},\;\underbrace{\sin 2\theta,\;\cos 2\theta}_{\text{orientation}}\big],
\end{equation}
reusing per-building quantities already defined: footprint area $A$ and height $z^{\uparrow}-z^{\downarrow}$; convexity $\operatorname{cvx}=A/A(\operatorname{conv})$ (the per-building form of $\phi_{\text{conc}}$); circularity $\mathrm{circ}=4\pi A/\ell^2$ (as in $\phi_{\text{circ}}$); elongation $\mathrm{elo}$ (as in $\phi_{\text{elo}}$); and the longest-edge orientation $\theta$ (as in $\phi_{\text{dir}}$, \mbox{Equation~(\ref{eq:dir})}), encoded as $(\sin 2\theta,\cos 2\theta)$ to fold the $180^{\circ}$ symmetry. This is the implemented \texttt{classic} feature subset of the Fr\'echet metric. We summarize the target buildings and the pooled context buildings as two diagonal Gaussians---per-feature means and standard deviations $(\mu_f^t,\sigma_f^t)$ and $(\mu_f^c,\sigma_f^c)$, computed after z-normalizing by the context---and take their diagonal Fr\'echet distance:
\begin{equation}
\label{eq:frechet}
\mathcal F(m\mid c_{1:Q})=\sum_f\left(\mu_f^t-\mu_f^c\right)^2+\left(\sigma_f^t-\sigma_f^c\right)^2,
\qquad
E(m)=e^{-\mathcal F}.
\end{equation}
A diagonal (rather than full) covariance is used because a single massing usually has too few buildings to estimate cross-feature correlations reliably. We evaluate this metric on two feature vectors. The multi-feature score $\mathrm{Fr}_{\mathrm{multi}}$ uses all seven features above---two or three representatives per characteristic family, rather than every score, since the full cross-product is left to the learned ensemble. The orientation score $\mathrm{Fr}_{\mathrm{orient}}$ keeps only the pair $(\sin 2\theta,\cos 2\theta)$; we single orientation out because the preliminary single-characteristic sweep already flags it as by far the strongest cue (Section~\ref{subsec:analysis}), so a metric devoted to the orientation distribution is worth its own score. This construction mirrors the FID idea of comparing two feature distributions using a Fr\'echet distance~\cite{heusel2017gans}, but~on interpretable massing geometry rather than network~embeddings.

\textbf{Orientation-distribution metric.}
For the same reason, we add one metric tailored entirely to orientation. Orientation is the clearest multi-peaked case---a block usually has one or two grid directions---so rather than reduce it to a scalar, we compare the target and context orientation distributions directly. We treat orientation as circular data~\cite{mardia2000directional} and fit a von~Mises kernel density estimate (KDE) to the context angles: the circular analog of a Gaussian KDE, where each observed angle is smoothed by a bell-shaped von~Mises kernel instead of a Gaussian~\cite{silverman1986density,mardia2000directional}. Let $\{2\theta_r^c\}_{r=1}^{R_c}$ be the doubled orientations of the $R_c$ buildings pooled over all context massings and $\{2\theta_r^m\}_{r=1}^{R_m}$ those of the $R_m$ target buildings (doubling maps the half-circle $[0,\pi)$ onto the full circle, encoding the $180^{\circ}$ symmetry). The~estimated density and the target score $\mathrm{KDE}_{\mathrm{orient}}$ are
\begin{equation}
\label{eq:kde}
\hat p(\alpha)=\frac{1}{R_c}\sum_{r=1}^{R_c}\frac{\exp\!\left(\kappa\cos(\alpha-2\theta_r^c)\right)}{2\pi I_0(\kappa)},
\qquad
E(m)=\exp\!\left(\frac{1}{R_m}\sum_{r=1}^{R_m}\log \hat p(2\theta_r^m)\right),
\end{equation}
where $I_0(\kappa)$ is the modified Bessel function of the first kind of order zero, which normalizes the von~Mises kernel. The~concentration $\kappa$ acts as an inverse KDE bandwidth (large $\kappa$~=~narrow kernel) and is set automatically from the circular spread of the context angles by a Silverman-type rule of thumb~\cite{silverman1986density}, then multiplied by the fixed factor used in the benchmark configuration: tightly clustered context orientations give a large $\kappa$ (sharp peaks), dispersed ones a small $\kappa$ (smooth density). The~score is the exponentiated average log-likelihood of the target angles under the context density; it is high when the target reuses the context's preferred directions and low~otherwise.

The three constructions above generate many more scores than we keep. The~fully designed pool contains about $85$ candidates: every characteristic under both reductions, the~two Fr\'echet variants, and~the orientation KDE, and~the broader set of generic complexity descriptors noted above (element counts and footprint/polygon shape ratios). We rank them all with the preliminary single-score sweep (Section~\ref{subsec:analysis}) and keep the $17$ scores of Table~\ref{tab:metric-inventory}: each characteristic under its better reduction (direction and height under both), plus the three distribution scores. The~discarded candidates score near chance (ROC--AUC below about $0.57$) and are strongly correlated with the scores we retain, so they would only add dimensionality and noise; we therefore do not use all designed features, and~even the largest learned set is this curated $17$, not the raw~pool.

\subsubsection{Learnable~Ensembles}
\label{subsubsec:learnable_ensembles}

\begin{figure}[H]

\includegraphics[width=\textwidth]{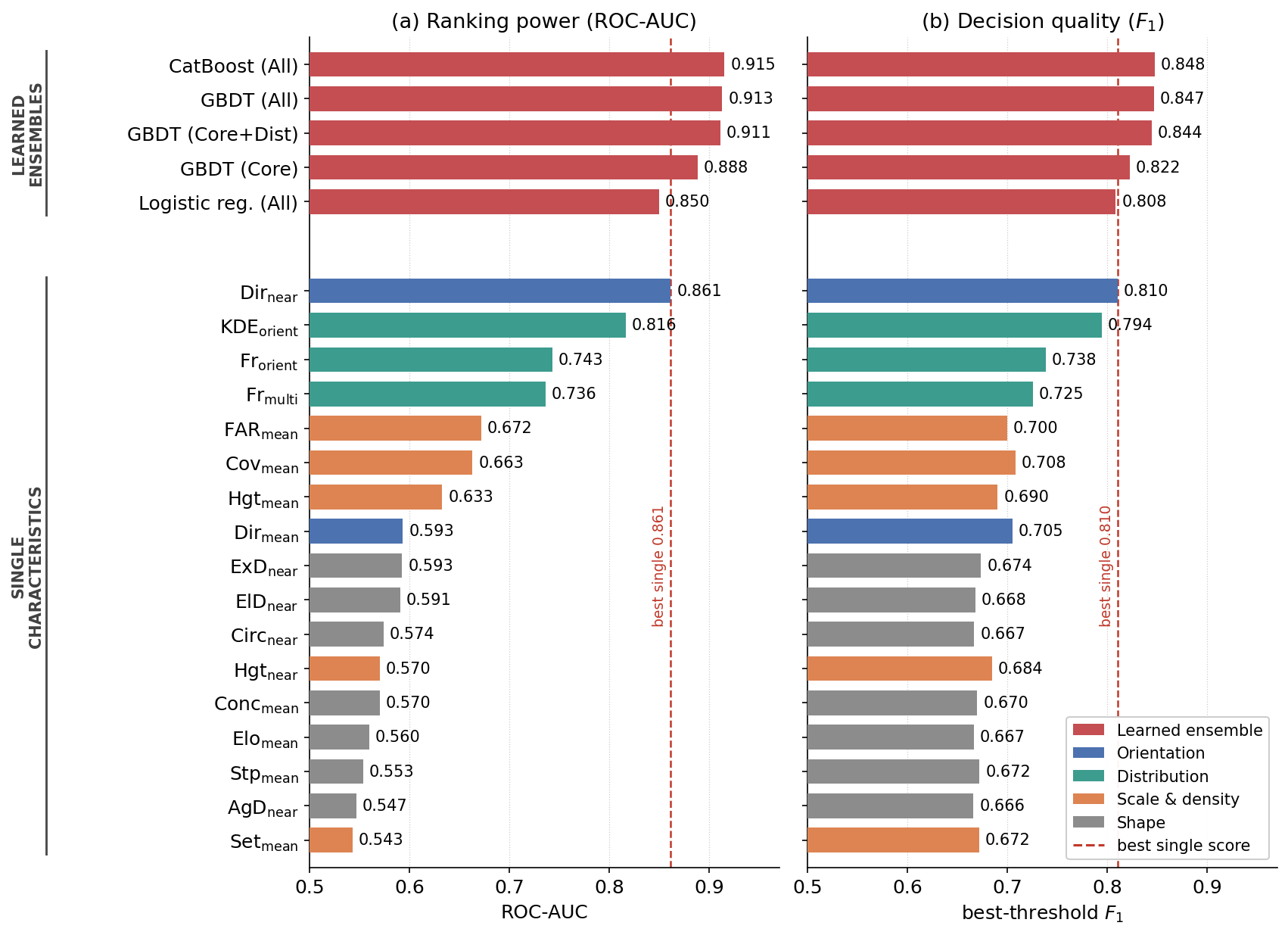}
\caption{Benchmark comparison of contextual-relevance metrics (symbols as in Table~\ref{tab:metric-inventory}). ROC--AUC ranks genuine above foreign contexts; $F_1$ measures best-threshold binary decisions. The~lower block reports the $17$ single scores, colored by characteristic family (orientation, scale \& density, shape) and by distribution metric; the upper block reports learned ensembles over the nested feature sets \textsc{Core}~($9$), \textsc{Core+Dist} ($12$), and~\textsc{All} ($17$) of Section~\ref{subsec:space}. Orientation scores dominate among single scores, and~the CatBoost ensemble over \textsc{All} gives the best overall~result.}
\label{fig:metric-overview}
\end{figure}

Each hand-designed score captures one facet and is individually imperfect. An~overall contextual-fit score could be a fixed weighted sum of per-facet scores~\cite{biljecki2022morphology}, but~the weights are hard to set by hand, so we learn the combination instead. To~do so, we first organize the scores above into three nested feature sets (Table~\ref{tab:metric-inventory}). \textsc{Core} holds the nine single-characteristic scalars---the best reduction of each of the nine strongest characteristics. \textsc{Core+Dist} adds the three distribution scores ($\mathrm{Fr}_{\mathrm{orient}},\mathrm{KDE}_{\mathrm{orient}},\mathrm{Fr}_{\mathrm{multi}}$), which are orientation-heavy and the most informative non-scalar cues. \textsc{All} adds five extra scalars: the alternate reductions $\mathrm{Dir}_{\mathrm{mean}},\mathrm{Hgt}_{\mathrm{near}}$ and the three weakest characteristics $\mathrm{Set}_{\mathrm{mean}},\mathrm{Stp}_{\mathrm{mean}},\mathrm{AgD}_{\mathrm{near}}$. The~split is by preliminary strength: the nine \textsc{Core} characteristics are the strongest single cues on the benchmark, whereas the extras are weak or redundant and are demoted to \textsc{All} only to test whether an ensemble can still extract residual signal from them (Section~\ref{subsec:analysis}, Figure~\ref{fig:metric-overview}).

Given a feature set, we stack the per-sample scores into a vector $\mathbf x(m,c_{1:Q})=(E_1,\dots,E_M)$ and fit a classifier $g_{\boldsymbol\beta}$ to predict the genuine/foreign label,
\begin{equation}
\hat y=g_{\boldsymbol\beta}\!\left(\mathbf x(m,c_{1:Q})\right)\in(0,1).
\end{equation}

We try three classifier families that span a deliberate range of capacity: logistic regression as a linear baseline, a~small gradient-boosted decision-tree (GBDT) model for fast non-linear feature-set screening, and~CatBoost~\cite{prokhorenkova2018catboost} as the higher-capacity final model. We do not run a full grid of every classifier on every feature set: the linear baseline and the fast GBDT are enough to screen which feature set is worth keeping, and~we then train the final metric with CatBoost on that set. A~tree ensemble is the natural choice because it captures the near-threshold response of orientation alignment together with additive contributions from the scale, density, and~shape families---interactions a single scalar score cannot represent. Every ensemble is trained and reported strictly under GroupKFold by massing ID, so its score reflects generalization to unseen sites rather than the paired structure of the negative-sampling~set.

\subsection{Metric Selection and~Analysis}
\label{subsec:analysis}

Figure~\ref{fig:metric-overview} summarizes the benchmark in two complementary views: ROC--AUC measures how well a score ranks genuine above foreign contexts, while the best-threshold $F_1$ measures the quality of a binary accept/reject decision. Read by characteristic family, the~picture is clear. Orientation is the dominant cue: $\mathrm{Dir}_{\mathrm{near}}$ alone reaches $0.861$ ROC--AUC, and~the two orientation distribution scores $\mathrm{KDE}_{\mathrm{orient}}$ ($0.816$) and $\mathrm{Fr}_{\mathrm{orient}}$ ($0.743$) are the next best, confirming that street-grid alignment carries most of the contextual signal. The scale \& density family follows at a distance---mean floor-area-ratio ($\mathrm{FAR}_{\mathrm{mean}}$), $\mathrm{Cov}_{\mathrm{mean}}$, and~$\mathrm{Hgt}_{\mathrm{mean}}$ sit around $0.63$--$0.67$---while the shape family (elongation, concavity, circularity, and~the diversities) is weakest, mostly between $0.55$ and $0.59$. Within~the orientation family, the reduction matters sharply: $\mathrm{Dir}_{\mathrm{near}}$ ($0.861$) far outperforms $\mathrm{Dir}_{\mathrm{mean}}$ ($0.593$), exactly as expected for a multi-peaked, grid-aligned quantity, and~the three orientation scores ($\mathrm{Dir}_{\mathrm{near}}$, $\mathrm{KDE}_{\mathrm{orient}}$, $\mathrm{Fr}_{\mathrm{orient}}$) each contribute somewhat differently, hinting at a non-linear dependence that a single score cannot~capture.

The strong contribution of orientation should not be regarded as a Melbourne-specific phenomenon, but~as a meaningful component of urban morphology. Building orientation is included among the morphological indicators used for comparative analysis of urban form at the individual-building and aggregate levels~\cite{biljecki2022morphology}. At~the street-network scale, orientation distributions have been shown to characterize spatial order across 100 cities, including regular grids, competing grids, and~irregular networks~\cite{boeing2019urban}. This interpretation is also consistent with approaches that treat buildings, parcels, and~roads as interdependent components of urban structure~\cite{domingo2019graph}. Orientation additionally has direct environmental consequences, particularly for solar access: facade and street-canyon orientation influence incident solar radiation and solar availability across different climatic and urban contexts~\cite{chatzipoulka2018sky,mohajeri2019solar}.

The weaker families are still useful because they add complementary information, which we confirm with the nested feature sets of Section~\ref{subsec:space}. A~GBDT on \textsc{Core} ($9$ scalars) already reaches $0.888$ ROC--AUC; adding the three distribution scores (\textsc{Core+Dist}, $12$) lifts it to $0.911$, the~single largest jump and further evidence that orientation distributions matter. The~best learned metric is CatBoost over \textsc{All} \cite{prokhorenkova2018catboost}, at~$0.915$ ROC--AUC and $0.848$ $F_1$ under five-fold GroupKFold on the $25\,153$ samples where all base scores are finite; the five extra scalars in \textsc{All} add only $0.004$ over \textsc{Core+Dist}, confirming that they were rightly demoted. Logistic regression on \textsc{All} stays at $0.850$---close to the best orientation-only score---so the gain of the tree models comes from non-linear interactions among the orientation, scale, and~density cues, which is why we screen feature sets with the fast GBDT but train the final metric with~CatBoost.

\begin{figure}[H]

\includegraphics[width=.9\textwidth]{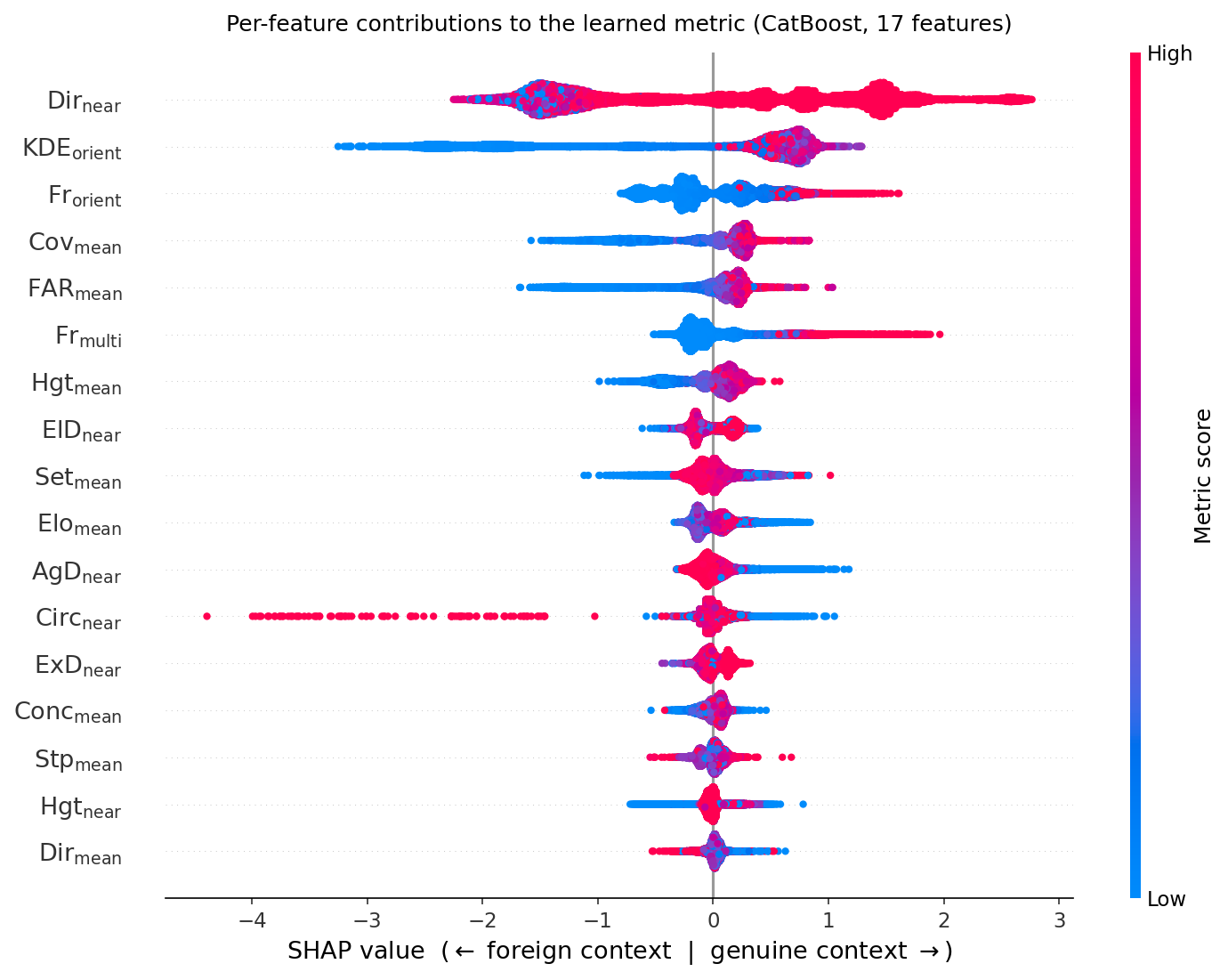}
\caption{TreeSHAP summary for the CatBoost metric over all $17$ scores of Table~\ref{tab:metric-inventory} (each row is one score, symbols as in the table). Scores are ordered top-to-bottom by mean attribution magnitude; each point is one benchmark sample, its horizontal position is the score's SHAP contribution (left~=~pushes towards foreign, right = towards genuine), and~its color is the underlying metric score (red high, blue low). Orientation scores provide the strongest and most directional evidence, while the extra scalars at the bottom contribute~little.}
\label{fig:shap}
\end{figure}

To check that the CatBoost score uses meaningful cues rather than artifacts, we inspect TreeSHAP attributions~\cite{lundberg2017shap}. Figure~\ref{fig:shap} shows that $\mathrm{Dir}_{\mathrm{near}}$ has the largest contribution by a wide margin, followed by $\mathrm{KDE}_{\mathrm{orient}}$ and $\mathrm{Fr}_{\mathrm{orient}}$---the same orientation family that dominated the single-score benchmark. The~scale \& density scores ($\mathrm{Cov}_{\mathrm{mean}},\mathrm{FAR}_{\mathrm{mean}},\mathrm{Hgt}_{\mathrm{mean}}$) contribute smaller corrections, while most shape and diversity scores have little influence. The~color gradient confirms the expected direction: high orientation-alignment scores push a pair towards genuine, and~low scores push it towards foreign.

\begin{figure}[H]

\subfloat[Genuine context.]{%
  \includegraphics[width=0.45\textwidth]{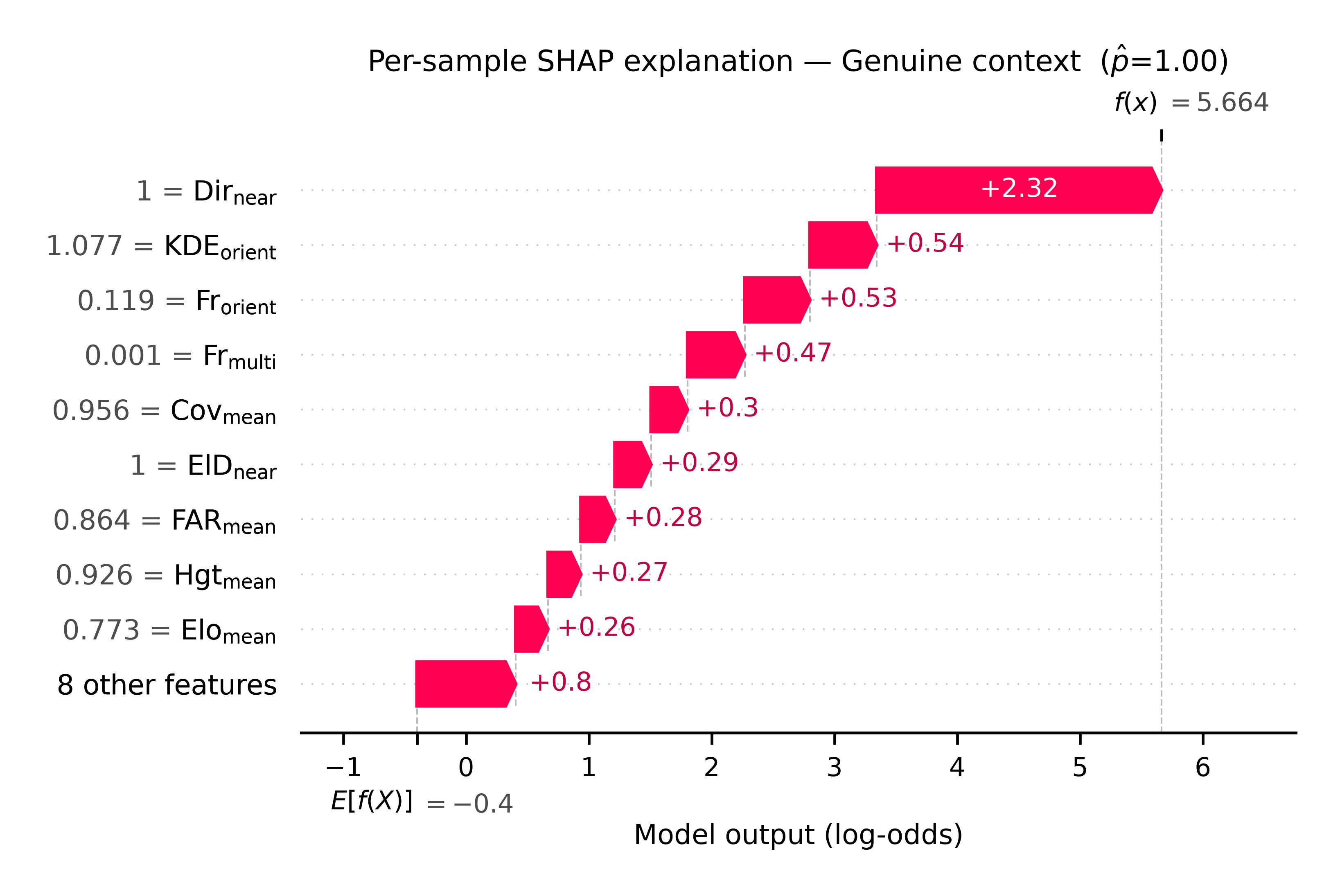}%
}\hfill
\subfloat[Foreign context.]{%
  \includegraphics[width=0.45\textwidth]{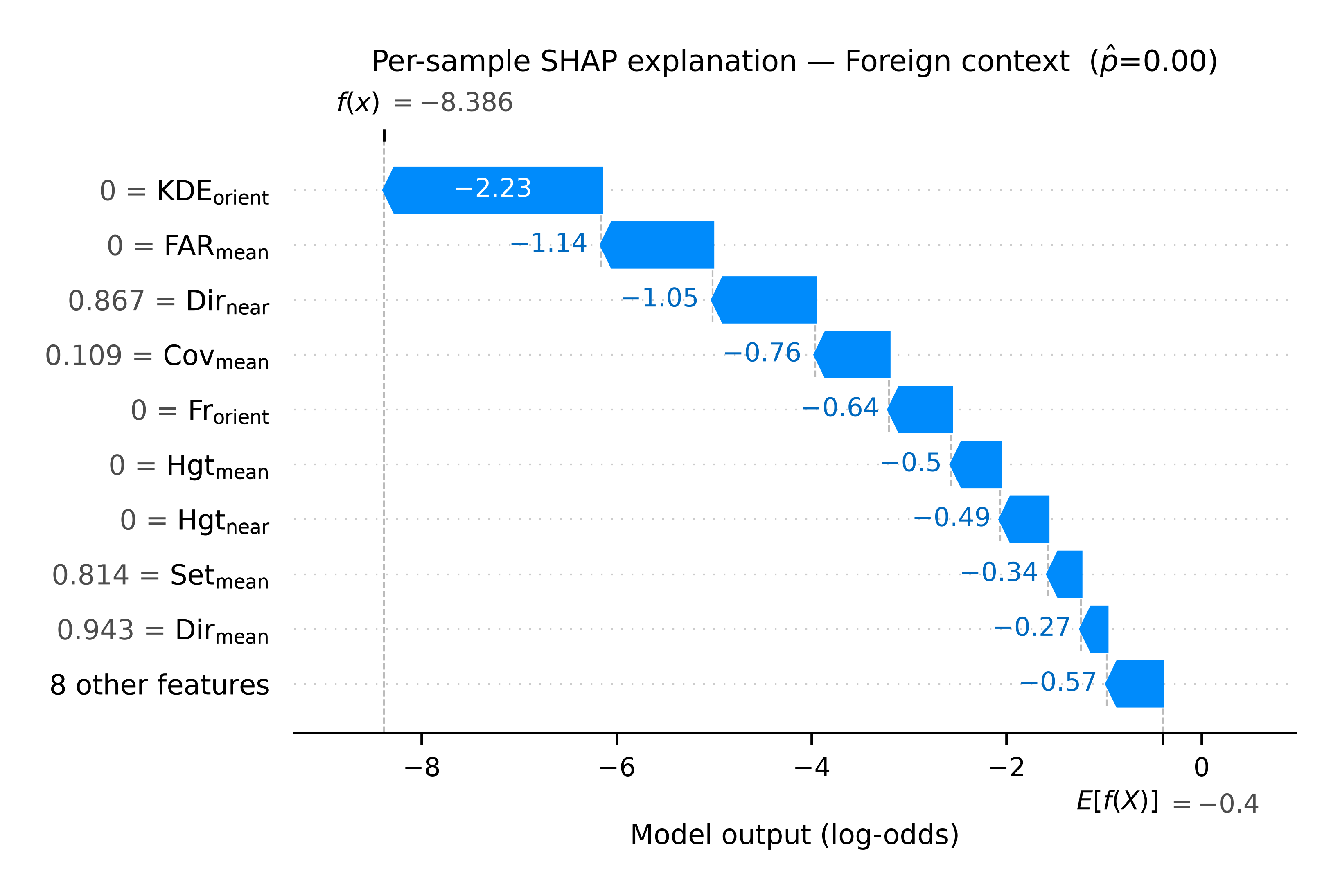}%
}
\caption{Two per-sample CatBoost explanations. The~accepted pair is driven mainly by orientation agreement; the rejected pair is driven mainly by orientation~mismatch.}
\label{fig:shap-waterfall}
\end{figure}

Figure~\ref{fig:shap-waterfall} gives two representative decisions. In~the genuine example, strong orientation agreement is the main positive contribution, with~density and scale adding smaller support. In~the foreign example, orientation mismatch dominates the rejection. These explanations show that the learned metric is interpretable: it primarily tests street-grid alignment, then refines the decision with density and scale~cues.

\vspace{-6pt}

We therefore use CatBoost over all $17$ contextual-relevance scores~\cite{prokhorenkova2018catboost} as the evaluation metric for the remaining experiments. It has the strongest benchmark performance, generalizes under grouped cross-validation, and~remains interpretable through its dominant orientation signal and secondary density/scale~refinements.

\section{Experiment~Setup}
\label{sec:experiments}

The task formulation introduced above allows us to study contextual massing generation using different representations of the surrounding urban context. In~this section, we specify the experimental setup used for this analysis. The~main goal is to answer two questions: whether combining context modalities during training helps a VLM better use each individual modality, and~whether combining modalities during inference improves generation quality compared with isolated context inputs. The~concrete context representations analyzed in this study are introduced in Section~\ref{subsec:modality_setup}.

We use the Qwen3-VL model family as the VLM backbone for this study, as~described in Section~\ref{subsec:models}. The~experimental pipeline consists of three steps. First, we prepare a dataset for training and inference, described in Section~\ref{subsec:dataset}. Second, we fine-tune the models under several context-presentation strategies, described in Section~\ref{subsec:training_setup}. Third, we evaluate the trained models across different context formats and levels of contextual information at inference time, as~described in Section~\ref{subsec:inference_setup}.

\subsection{Models}
\label{subsec:models}

We use the Qwen3-VL model family as the base architecture for all experiments~\cite{qwen3vl}. This choice is motivated by the fact that the Qwen series is an actively maintained open-source model family and has demonstrated strong performance across diverse downstream tasks. Although~Qwen3-VL is not the newest available VLM, it provides a stable and established backbone for controlled experiments with multimodal context. Qwen3-VL is designed to process interleaved text, image, and~video inputs while producing textual outputs. The~family includes dense variants of different scales; in this work, we use the 2B, 4B, and~8B parameter models to analyze how model capacity affects contextual massing~generation.

Qwen3-VL follows a three-module architecture consisting of a vision encoder, a~vision-language merger, and~a Qwen3 large language model backbone. The~vision encoder is based on SigLIP-2~\cite{siglip} and supports dynamic native-resolution visual inputs, mapping images to variable-length visual tokens. These tokens are then passed through an MLP-based vision-language merger, which compresses local groups of visual features and projects them into the language model's hidden dimension. The~resulting visual tokens are processed jointly with textual tokens by the Qwen3 decoder, which autoregressively generates the output~sequence.

\subsection{Dataset}
\label{subsec:dataset}

To conduct the experiments, we assembled a multimodal dataset of building massings from Melbourne, Australia. Each record in the resulting dataset corresponds to one target massing and contains the site contour, the~three-dimensional geometry of buildings located on the site according to the task definition in Section~\ref{sec:task}, the~absolute position of the local coordinate system, and~the identifiers of neighboring massings located within a 100-m radius of the target massing. These neighboring massings are later used to construct the vector and visual context representations described in Section~\ref{subsec:modality_setup}. Thus, the~dataset serves as the empirical basis for studying how VLMs use urban context in structured massing generation and for training and testing the proposed contextual compatibility~metrics.

The dataset was built from source data obtained through the City of Melbourne Open Data portal~\cite{melbourne}. This portal provides public access to more than 200 datasets maintained by the City of Melbourne, covering urban infrastructure, buildings, mobility, public space, and~other municipal data. In~our pipeline, these public tabular datasets were used as the raw information source for site geometry and building massing geometry. The~specific tables and fields used for each data component are summarized in Table~\ref{tab:data_sources}. We then transformed, aligned, and~merged these source data into the unified massing representation used in this work. The~implementation of the dataset construction algorithm is provided in the Supplementary~Materials.

\begin{table}[H]
  \caption{Datasets sourced from the City of Melbourne Open Data portal~\cite{melbourne} used to construct an experimental set for VLM training and~inference.}
  \label{tab:data_sources}
  \centering
  \small
  \begin{tabularx}{\textwidth}{p{4.2cm} X}
      \toprule
      \textbf{Dataset Id} & \textbf{Description} \\
      \midrule
      \texttt{property-boundaries} & The boundaries of all properties within the City of Melbourne. A~boundary is described as a division between adjacent political entities, tracts of private land, or~geographic zones. This dataset is a source of site~contours. \\
      \midrule
      \texttt{2023-building-footprints} & The footprints of all structures within the City of Melbourne. A~building footprint is a 2D polygon (or multi-polygon) representation of the base of a building or structure. The~footprint is defined as the boundary of the structure where the walls intersect with the ground plane or podium, rather than an outline of the roof area (roofprint). Where a building undergoes a significant change in built form, multiple footprint polygons are ‘stacked’ vertically to define the built form. This dataset provides 3D building~geometry. \\
      \bottomrule
  \end{tabularx}
\end{table}

After preprocessing and filtering, the~final dataset contains 12,845 target massing samples. For~the VLM experiments, we randomly split the dataset into 10,276 training samples and 2569 test samples. For~each target massing, we additionally constructed a neighborhood index by selecting all massings from the final dataset whose geometry lies within a 100-m radius of the target site. Since the dataset represents a single city, a~random split may place geographically adjacent sites in different subsets, potentially resulting in partial overlap between their surrounding urban contexts. We analyze the extent of this overlap and its effect on the reported evaluation results in Appendix~\ref{app:context_overlap}.

Site Contour is represented as a JSON structure specifying a list of polygons. Each polygon is defined by a list of points, and~each point is a list of two real numbers rounded to two decimal places specifying the X and Y coordinates in meters in a local Cartesian coordinate system. One of the contour points is taken as the origin of this coordinate system. The~structure does not explicitly distinguish between exterior and interior polygons; however, during~dataset construction, geometric validity is guaranteed: polygons intersect only when one is completely inside the other. In~addition, if~a site is defined by more than one polygon, one of them is external and all others are internal, so the polygons form one geometrically connected~structure.

Massing Geometry describes all buildings located on the target site. We use LoD 1.3 building geometry~\cite{biljecki2016improved}, represented as a JSON structure following the format introduced in Section~\ref{sec:task} and illustrated in Figure~\ref{fig:json_vector_representation}. In~this representation, each building is decomposed into elementary extrusions with polygonal bases and bottom and top elevations. This keeps the output structured, editable, and~directly measurable while remaining suitable for text-based generation by~VLMs.

Global Position links each local massing coordinate system to a global metric GIS reference frame. The~site contour and target massing geometry are stored, provided to the model, and~generated in relative local coordinates. For~each sample, we additionally store the absolute coordinates of the local origin in EPSG:3857. Since the first point of the site contour is used as the zero point of the local coordinate system, this global base point defines the absolute position of the entire site contour and massing geometry. The~same relative-coordinate convention is used when neighboring massings are provided as 1D context. Absolute coordinates are therefore not part of the generation target; they are used only to align the target and context massings for visualization, including creating 3D context views and rendering generated massings within their surrounding urban~context.

Surrounding Context is defined by neighboring massings from the dataset. For~each target massing, we store the identifiers of all massings located within a 100-m radius. These identifiers define the local urban context used for metric computation and for constructing the context representations used in VLM experiments. The~concrete vector and visual representations derived from this neighborhood are described in Section~\ref{subsec:modality_setup}.

\subsection{Modality~Setup}
\label{subsec:modality_setup}

We consider three representations of surrounding urban context: 1D textual context, 2D map context, and~3D visual context. These formats are selected to analyze how VLMs use vector-like and image-based representations of three-dimensional urban objects and to assess whether these representations provide complementary information. The~1D context uses the same structured geometry format as the generated output, while the visual contexts provide two different image-based views of the same environment: a top-down 2D representation and perspective 3D views. This allows us to test whether the way three-dimensional objects are projected into images affects the information that a VLM can extract from~them.

As described in Section~\ref{subsec:dataset}, the~surrounding context for each target massing is defined as the set of dataset massings located within a 100-m radius. The~concrete number of context elements used during training and inference is controlled by the sampling procedures described in Sections~\ref{subsec:training_setup} and~\ref{subsec:inference_setup}.

The 1D context represents neighboring buildings as textual JSON geometry. For~each sample, we select from 1 to 10 massings from the surrounding context and serialize their building geometry using the same hierarchical representation as the target massing. The~final input object has the same general structure as the massing format shown in Figure~\ref{fig:json_vector_representation}, but~its buildings correspond to neighboring context buildings rather than to buildings on the target site. Thus, the~1D context provides the model with editable vector-like geometric information in textual~form.

The 2D context is a single top-down map image of the area around the development site. We use an OpenStreetMap-style map rendering, in~which the target site is highlighted as a red-filled polygon with a black outline. This representation captures the horizontal spatial organization of the surrounding area, including the site's position relative to nearby streets and urban blocks. However, it does not provide vertical information about neighboring~buildings.

The 3D context consists of rendered perspective views of the target site and surrounding massing geometry. To~create these views, we convert the JSON representations of the context massings and the target site contour into meshes. We then render eight views from uniformly distributed horizontal angles around the site, covering the full $0^\circ$--$360^\circ$ range with a $45^\circ$ step. The~vertical camera angle is fixed at $60^\circ$. In~the renderings, the~target site is shown as a red filled area, while surrounding buildings are shown as gray volumes. The~number of 3D views provided to the model during training and inference is controlled in Sections~\ref{subsec:training_setup} and~\ref{subsec:inference_setup}.

\begin{figure}[H]
    \includegraphics[width=\textwidth, trim=40 230 40 245, clip]{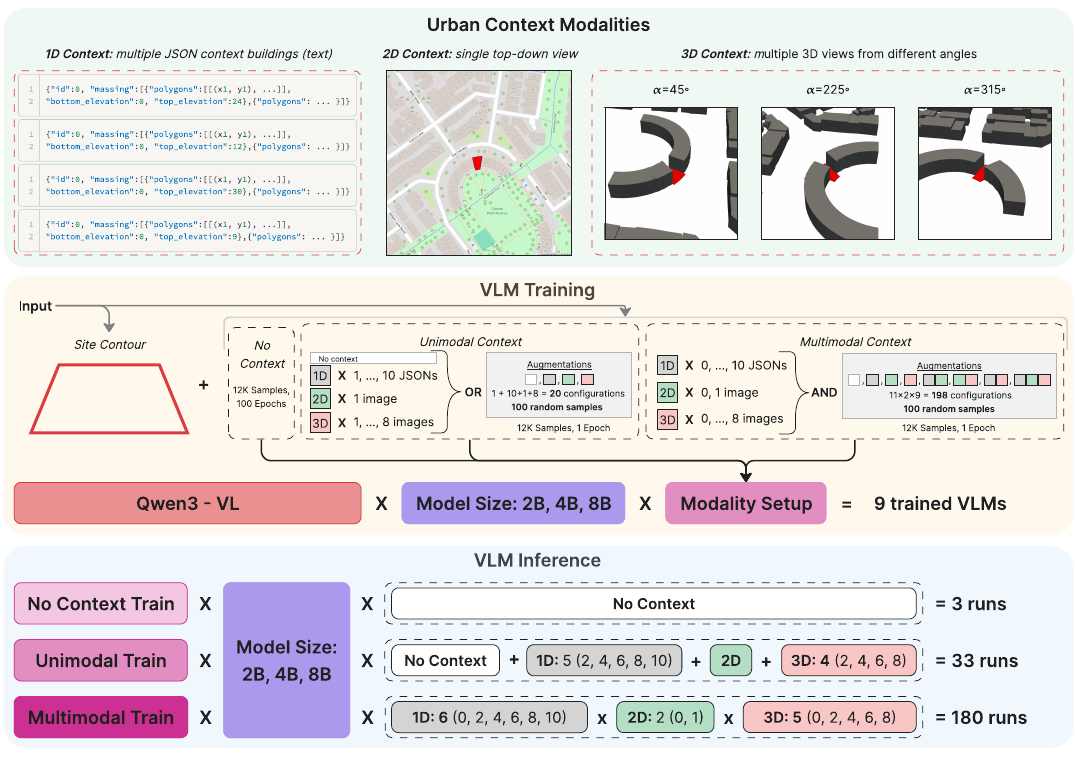}
    \caption{Experimental setup for training and inference. The~top block shows the three urban-context modalities used in the study: 1D textual JSON descriptions of neighboring massings, a~2D top-down map view, and~3D perspective views from multiple angles. During~training, Qwen3-VL models of three sizes are fine-tuned under no-context, unimodal-context, and~multimodal-context regimes. During~inference, the~trained models are evaluated under controlled context configurations to analyze the effects of modality type, modality combinations, and~the amount of contextual~information.}
    \label{fig:experiment_setup}
\end{figure}

Examples of all three context representations are shown in Figure~\ref{fig:experiment_setup}.

\subsection{Training~Setup}
\label{subsec:training_setup}

The training setup is designed to determine whether VLMs can learn to use each context modality more effectively when trained on samples that combine multiple modalities. The~training pipeline is summarized in Figure~\ref{fig:experiment_setup}. We use supervised fine-tuning~\cite{sft}: the model receives the site contour and, depending on the training regime, surrounding context as input, and~is trained to generate the target massing geometry. The~site-contour and massing formats follow the task formulation in Section~\ref{sec:task}.

To define different training regimes, we use a stochastic context-augmentation algorithm. For~each sample in the training split described in Section~\ref{subsec:dataset}, we generate 100~augmented variants. Each augmented instance is created by sampling whether context is included, which context modalities are present, and~how much information is provided for each selected~modality.

Let
\[
\mathcal{M}=\{\mathrm{1D},\mathrm{2D},\mathrm{3D}\}
\]
be the set of available context modalities. We first sample whether any surrounding context is included:
\[
C \sim \operatorname{Bernoulli}(p_c),
\]
where \(C=0\) means that the instance is presented without context and \(C=1\) means that context modalities can be included. If~\(C=1\), each modality is independently selected:
\[
B_m \sim \operatorname{Bernoulli}(p_m), \qquad m\in\mathcal{M},
\]
and the preliminary modality set is
\[
\mathcal{S}_0=\{m\in\mathcal{M}: B_m=1\}.
\]

No resampling is applied if \(\mathcal{S}_0=\varnothing\); in this case, the~augmented instance remains without~context.

The final modality set \(\mathcal{S}\) is controlled by the maximum number of modalities \(k_{\max}\). If~more than \(k_{\max}\) modalities are selected, we uniformly sample \(k_{\max}\) modalities from \(\mathcal{S}_0\):
\[
\mathcal{S}=
\begin{cases}
\varnothing, & C=0 \ \text{or}\ \mathcal{S}_0=\varnothing,\\
\mathcal{S}_0, & |\mathcal{S}_0|\leq k_{\max},\\
\operatorname{UniformSubset}(\mathcal{S}_0,k_{\max}), & |\mathcal{S}_0|>k_{\max}.
\end{cases}
\]

After \(\mathcal{S}\) is fixed, the~amount of context is sampled for modalities with variable size:
\[
n_{\mathrm{1D}} \sim \operatorname{Uniform}\{1,\dots,10\}, 
\qquad
n_{\mathrm{3D}} \sim \operatorname{Uniform}\{1,\dots,8\}.
\]

These counts are sampled only when the corresponding modality is included in \(\mathcal{S}\). The~2D modality always consists of one top-down~image.

Using this algorithm, we define three training regimes summarized in Table~\ref{tab:training_regimes}: no-context, unimodal-context, and~multimodal-context. The no-context regime serves as a baseline for massing generation from the site contour alone. Here, context is never included, which is equivalent to setting \(p_c=0\). Since all 100 augmented variants are identical with respect to the input context, this regime corresponds to training for 100 epochs on the dataset without surrounding-context information. This baseline estimates how well a model can produce massings that accidentally match the surrounding context without observing~it.

\begin{table}[H]
    \caption{Summary of training regimes used for context augmentation. $p_c$ is the probability that an augmented sample can contain surrounding context, $k_{\max}$ is the maximum number of context modalities in one sample, $N_{\mathrm{cfg}}$ is the number of unique context configurations per original site--massing pair, and~$E_{\mathrm{eff}}$ is the effective number of passes over these configurations under 100 augmentations per~sample.}
    \label{tab:training_regimes}
    \centering
    \begin{tabularx}{\textwidth}{wl{3.5cm} wl{2.2cm} wl{2.2cm} X wl{2.5cm}}
        \toprule
        \textbf{Regime} & $\boldsymbol{p_c}$ & $\boldsymbol{k_{\max}}$ & $\boldsymbol{N_{\mathrm{cfg}}}$ & $\boldsymbol{E_{\mathrm{eff}}}$ \\
        \midrule
        No context 
        & 0.0 
        & 0 
        & 1 
        & 100 \\
        Unimodal context 
        & 0.9 
        & 1 
        & $20$ 
        & $5$ \\
        Multimodal context 
        & 0.9 
        & 3 
        & $198$ 
        & $\approx$0.5 \\
        \bottomrule
    \end{tabularx}
\end{table}

The unimodal-context regime is designed to train models on isolated context modalities. We set \(p_c=0.9\), \(p_m=0.5\) for each modality, and~\(k_{\max}=1\). Thus, each augmented sample contains at most one modality: either 1D, 2D, 3D, or~no context. Since modality selection can also produce an empty set, the~final probability of a no-context sample is
\[
0.1 + 0.9 \cdot 0.5^3 = 0.2125.
\]

For each target pair of site contour and massing, the~augmentation procedure creates 100 variants with randomly selected modality type and context amount. There are \(1+10+1+8=20\) possible unimodal context configurations, so configurations are repeated during training; on average, this corresponds to approximately five passes over the unimodally augmented configuration~space.

The multimodal-context regime is designed to train models on combinations of context modalities. We use the same probabilities, \(p_c=0.9\) and \(p_m=0.5\), but~set \(k_{\max}=3\), allowing any subset of the three modalities to appear in the same sample. With~11 possible 1D states including absence, 2 possible 2D states including absence, and~9 possible 3D states including absence, this gives \(11 \times 2 \times 9 = 198\) possible context configurations. Since we still generate 100 augmented variants per original sample, repeated configurations are possible but less frequent; on average, training covers approximately half of the multimodally augmented configuration~space.

All images are resized to $256 \times 256$ pixels before training. We fine-tune the Qwen3-VL models using LoRA~\cite{lora}, updating only the attention projection modules \texttt{q\_proj}, \texttt{k\_proj}, \texttt{v\_proj}, and~\texttt{o\_proj}. The~LoRA rank is set to 16, $\alpha$ to 32, and~dropout to 0.05, with~no bias adaptation. We use the Adam optimizer~\cite{adam} with a learning rate of $10^{-4}$, batch size 1024, and~a cosine learning-rate schedule with a warmup ratio of 0.03~\cite{warmup}. The~same optimization settings are used for all model sizes and training~regimes.

\subsection{Inference~Setup}
\label{subsec:inference_setup}

The inference setup is designed to address the two research questions introduced above. First, we test whether multimodal training enables the model to use each modality more effectively. Second, we test whether combining modalities at inference time improves contextual massing generation compared with using each modality in isolation. All inference experiments are conducted on the test split of the dataset described in Section~\ref{subsec:dataset}. The~input and output formats follow the task formulation in Section~\ref{sec:task}: the model receives the site contour and optional context, and~generates the target massing~geometry.

Models trained in the no-context regime are evaluated only without surrounding context. This provides the baseline for massing generation conditioned solely on the target site~contour.

To answer the first question, we perform unimodal inference for both unimodally and multimodally trained models. Each model is evaluated separately with one context modality at a time: no context, 1D context, 2D context, or~3D context. This allows us to compare whether the same isolated modality is used differently across training strategies. We also vary the amount of context to analyze whether additional context elements improve generation quality and whether this effect changes after multimodal training. To~reduce the number of inference runs while preserving this trend analysis, we use every second context count from the training range: for 1D context, 2, 4, 6, 8, and~10 neighboring massing JSONs; for 3D context, 2, 4, 6, and~8 rendered views. The~2D context consists of one top-down image and therefore has a single~configuration.

To answer the second question, we perform multimodal inference with combinations of context modalities. This evaluation applies only to multimodal models, since they are the only models trained with modality combinations. We enumerate all combinations of 1D, 2D, and~3D context, including the no-context case, and~vary the amount of 1D and 3D context using the same reduced count grids: \(\{2,4,6,8,10\}\) for 1D and \(\{2,4,6,8\}\) for 3D. This setup allows us to compare single- and combined-modality inference for the same model and analyze whether different context representations provide complementary information about the surrounding urban~environment.

Counting runs by trained model type, the~unimodally trained models require
\[
(1 + 5 + 1 + 4) \times 3 = 33
\]
inference runs across the three model sizes. For~multimodally trained models, unimodal configurations are included as a subset of the full multimodal grid, giving
\[
(6 \times 2 \times 5) \times 3 = 180
\]
runs across the three model sizes. The~overall training and inference protocol is summarized in Figure~\ref{fig:experiment_setup}.

All image inputs are resized to \(256 \times 256\) pixels. Inference is performed using the vLLM~\cite{vllm} package. We sample outputs with temperature 1, top-\(p=1\), top-\(k=40\), and~presence penalty~2.

\section{Results}
\unskip

\subsection{Unimodal Inference~Analysis}

This section analyzes models trained under the regimes described in Section~\ref{subsec:training_setup} and evaluated in the unimodal inference setting described in Section~\ref{subsec:inference_setup}. The~goal is to answer whether combining modalities during training helps VLMs better use each individual modality at inference time. In~this setting, each model receives one context modality at a time: 1D, 2D, or~3D.

\begin{figure}[H]

    \includegraphics[width=\textwidth]{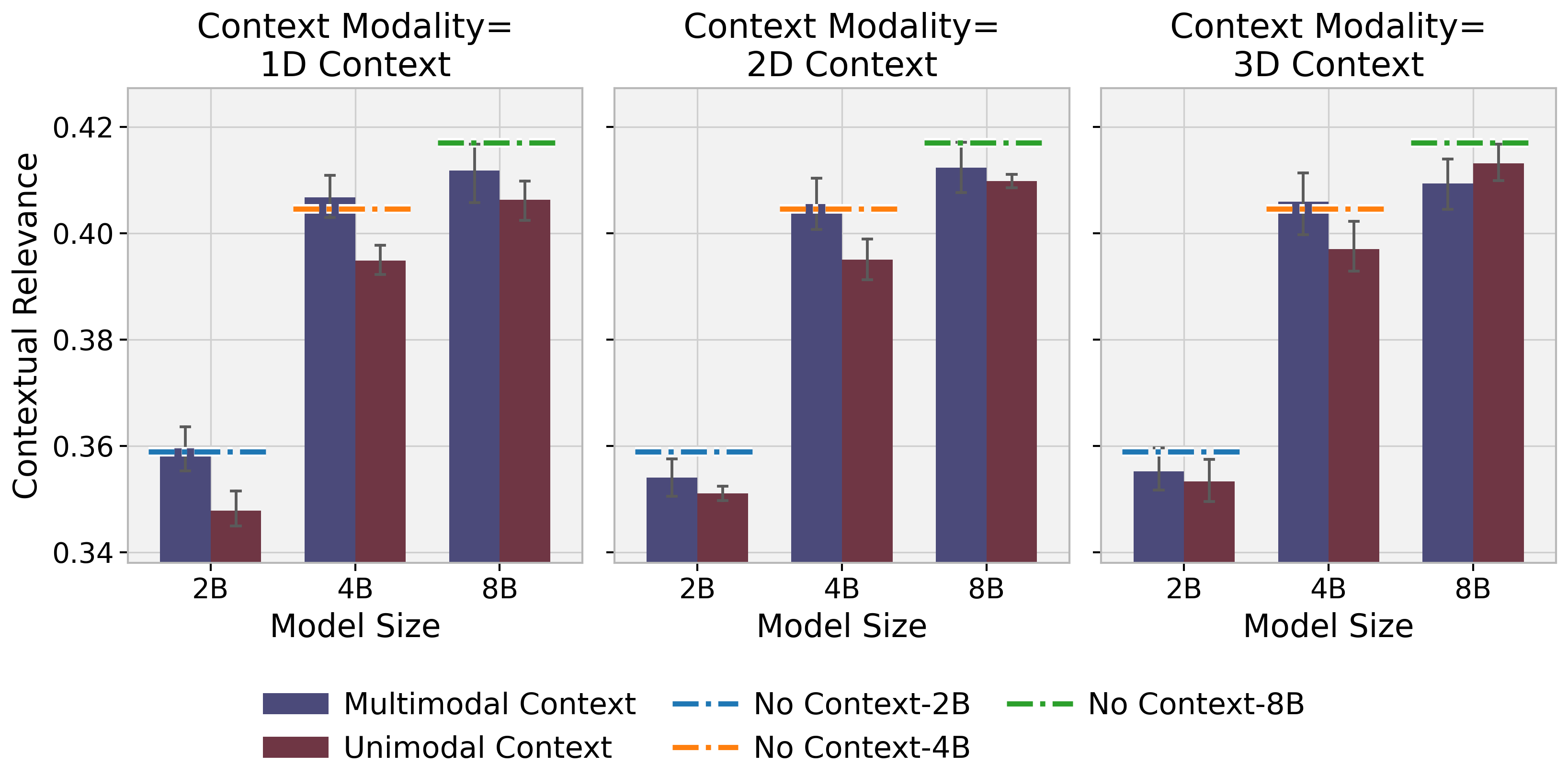}
    \caption{\textls[-15]{Contextual relevance score of unimodal inference configurations averaged over context counts. Bars show results for models trained with unimodal and multimodal context and evaluated with a single context modality. Results are grouped by inference modality and model size. Dashed horizontal lines show the scores of no-context baselines trained without surrounding-context~information.}}
    \label{fig:crossmodal_bar}
\end{figure}

We first compare modality use through the contextual relevance score averaged over the evaluated context counts. Figure~\ref{fig:crossmodal_bar} reports this average score for each inference modality, training strategy, and~model size. The~main pattern is that multimodally trained models outperform unimodally trained models in most single-modality inference configurations. This means that joint exposure to several context representations during training improves the model's ability to use an isolated modality at inference time. A~likely explanation is that multimodal training helps align structured and visual representations of the same urban environment, making each modality more informative even when it is provided~alone.

\begin{figure}[H]

    \includegraphics[width=\textwidth]{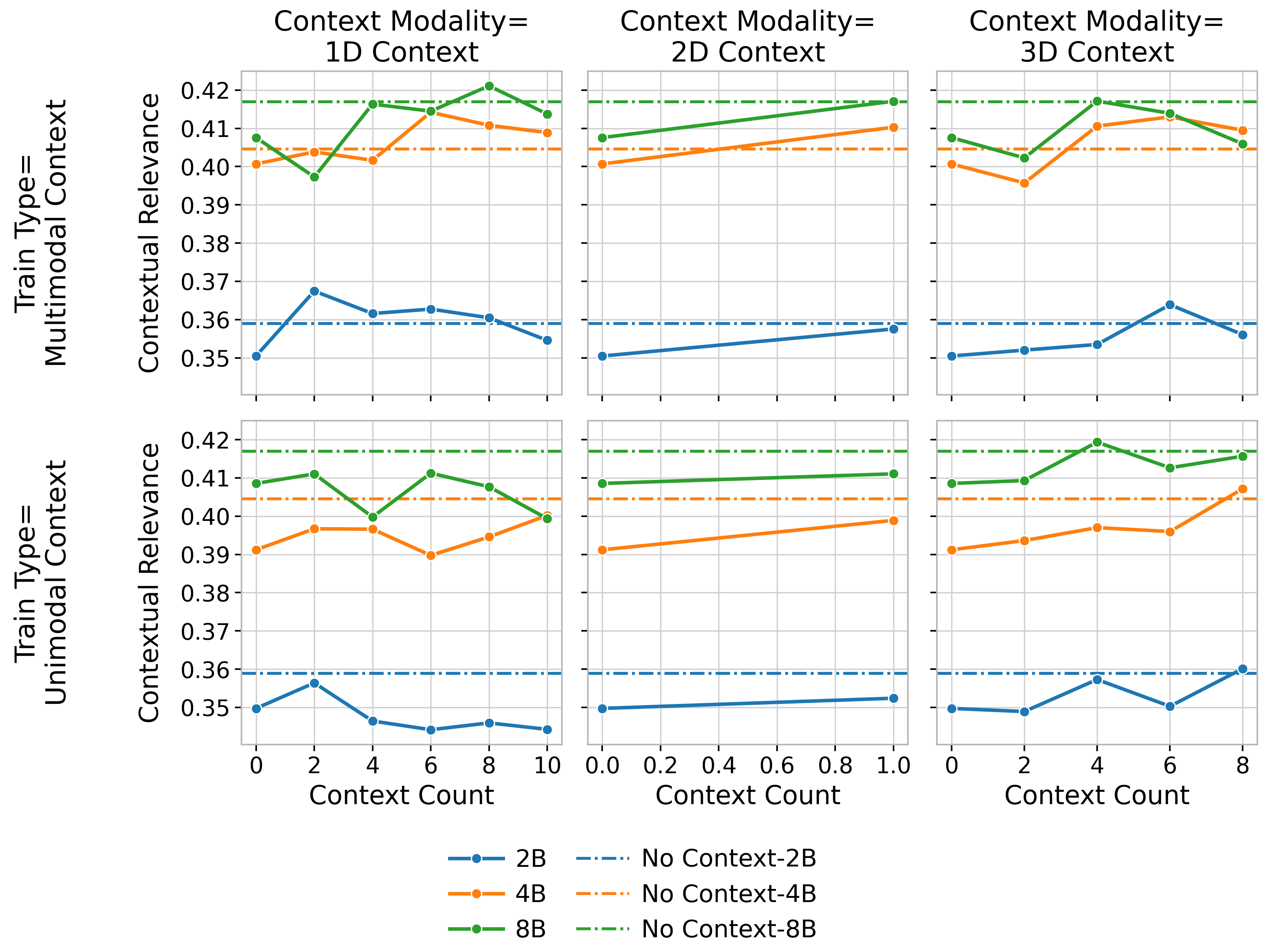}
    \caption{Effect of context amount in unimodal inference. Lines show the dynamics of contextual relevance score by context count for unimodally and multimodally trained models evaluated with a single context modality. For~1D context, the~x-axis denotes the number of neighboring massing JSONs; for 3D context, the~number of rendered views; for 2D context, the~x-axis shows the absence or presence of the top-down~image.}
    \label{fig:crossmodal_line_grid}
\end{figure}

We then analyze modality use through the dynamics of contextual relevance as the amount of context changes. Figure~\ref{fig:crossmodal_line_grid} shows the score for different numbers of 1D neighboring massings and 3D rendered views, while the 2D case corresponds to the presence or absence of a top-down view. In~most configurations, increasing the amount of context improves the score, although~the trend is not always monotonic. The~effect is stronger for multimodally trained models: adding more 1D or 3D context usually produces a clearer gain than in unimodally trained models. This supports the same conclusion from another angle: multimodal training not only improves average single-modality performance, but~also helps the model extract additional useful information when more context elements are~provided.

\unskip

Both figures also show a strong dependence on model size. For~the same modality and training strategy, larger models achieve higher contextual relevance averaged by context count. However, the~gain from scaling is sublinear: the improvement from 2B to 4B is generally larger than the improvement from 4B to 8B. This suggests that additional capacity helps VLMs process structured and visual context, but~the marginal benefit decreases within the evaluated model~range.

Finally, the~no-context baseline remains competitive in several configurations, especially for smaller models. This should be interpreted with caution. The~no-context models were trained for more effective epochs on the same site--massing pairs, while context-aware models had to learn from stochastic context augmentations. Thus, the~baseline may benefit from stronger memorization of dataset-level regularities, and~longer training of context-aware models could further improve their~results.

\subsection{Multimodal Inference~Analysis}

This section analyzes multimodally trained models under the multimodal inference setting described in Section~\ref{subsec:inference_setup}. The~goal is to answer whether adding contextual information through additional modalities at inference time improves context-aware massing generation. We therefore compare the same models when they receive a single context modality and when they receive combinations of 1D, 2D, and~3D~context.

\begin{figure}[H]

    \includegraphics[width=\textwidth]{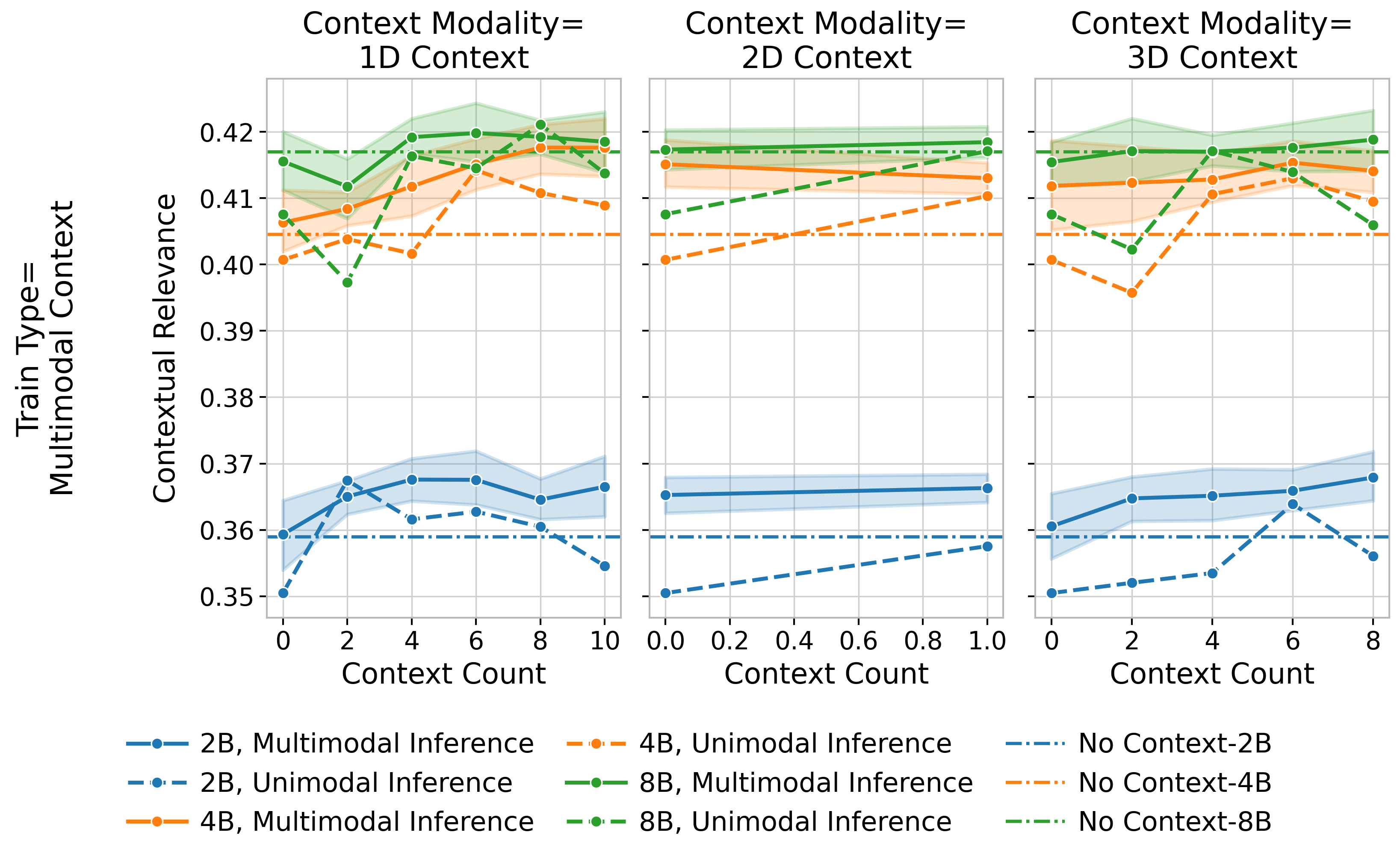}
    \caption{Comparison of unimodal and multimodal inference for each context modality. Solid lines show multimodal inference configurations averaged over the amounts of the other context modalities and plotted with 95\% confidence intervals. Short-dashed lines show inference with only the corresponding modality, while long-dashed lines show no-context baselines trained without surrounding-context~information.}
    \label{fig:multimodal_line_grid}
\end{figure}

We first evaluate the average effect of adding modalities. Figure~\ref{fig:multimodal_line_grid} compares single-modality inference with multimodal inference for each context type. For~each modality, the~single-modality line shows performance when only that modality is provided, while the multimodal line averages over configurations where the same modality is combined with other available modalities. Across modalities and model sizes, multimodal inference achieves higher contextual relevance than the corresponding single-modality inference. This indicates that additional context modalities provide useful information rather than only duplicating the signal already present in one~representation.

\begin{figure}[H]

    \includegraphics[width=\textwidth]{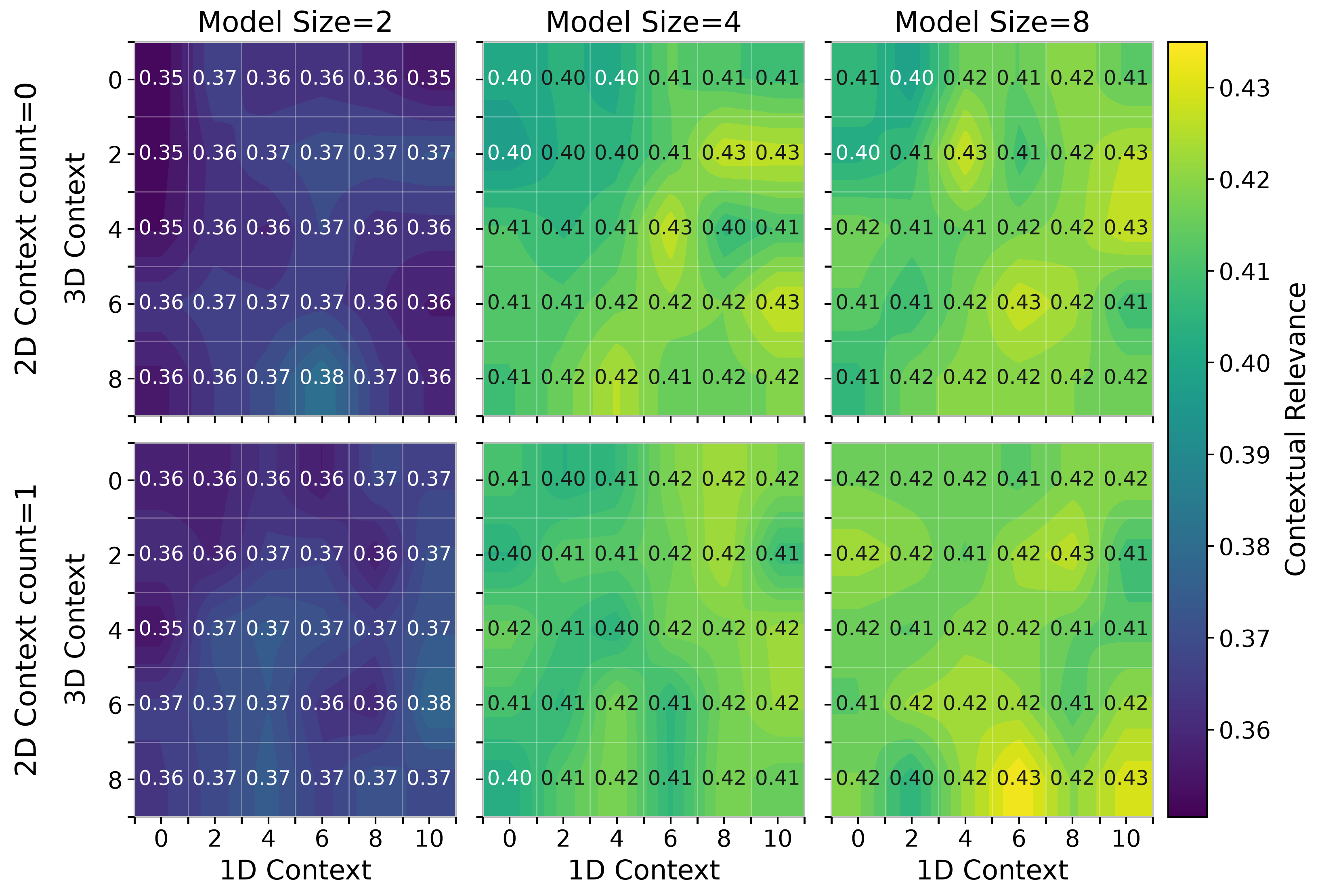}
    \caption{Multimodal context inference results for multimodally trained models. Heatmaps show contextual relevance as a function of the number of 1D neighboring massing JSONs and 3D rendered views, separately for different model sizes and for inference without or with the 2D top-down context~image.}
    \label{fig:multimodal_contour_grid}
\end{figure}

We then analyze how specific modality combinations affect contextual relevance. Figure~\ref{fig:multimodal_contour_grid} shows the score as a function of the number of 1D neighboring massing JSONs and 3D rendered views, separately for different model sizes and for inference with or without the 2D top-down image. The~joint increase in 1D and 3D context generally improves the score across model sizes. This suggests that textual geometric descriptions and perspective visual views capture complementary properties of the surrounding urban fabric: the former provides structured massing geometry, while the latter exposes the model to visual relationships of scale, volume, and~spatial~arrangement.

The effect of adding 2D context is more localized. The~top-down view improves performance mainly when the amount of 1D and 3D context is small. In~these cases, it provides useful information about horizontal spatial organization, such as site position, footprint arrangement, and~the nearby urban block structure. When many 1D massings and 3D views are already available, the~additional gain from the 2D image becomes smaller. This suggests that part of the information in the top-down map is covered by richer structured and perspective context, while its contribution is most useful when other context sources are~limited.

Overall, multimodal inference improves contextual massing generation because different context representations contribute complementary information. The~results show that, for~VLMs, vector geometry, top-down images, and~3D perspective views are not interchangeable encodings of the same context; their combination changes the quality of generated massings and provides a stronger contextual signal than isolated~modalities.

\unskip

\subsection{Qualitative Analysis of Generated~Massings}

\begin{figure}[H]

    \includegraphics[width=\textwidth]{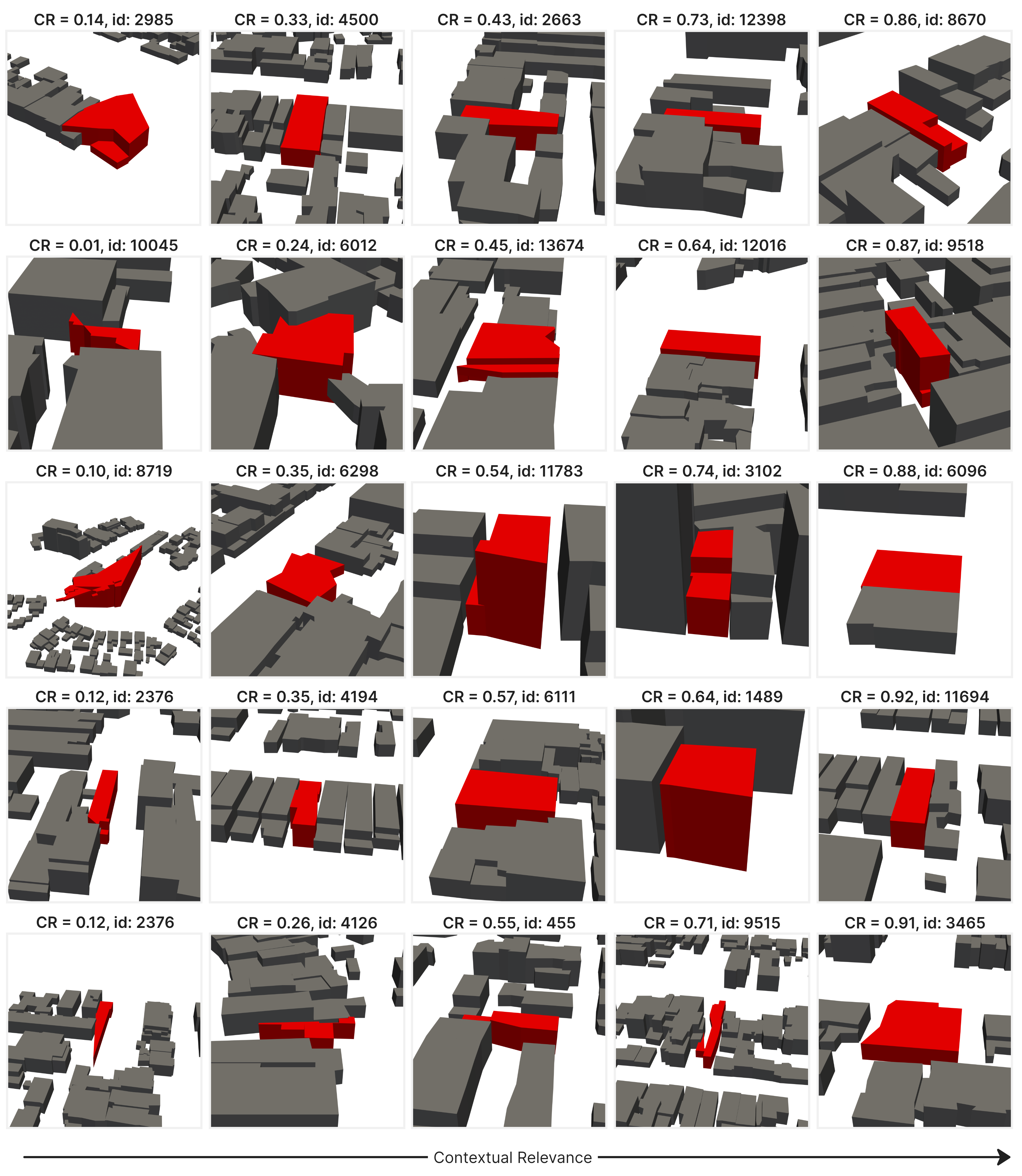}
    \caption{Qualitative examples of generated massings with different contextual relevance scores. Generated massings are shown in red, and~surrounding buildings are shown in gray. Examples are arranged from lower to higher contextual~relevance.}
    \label{fig:generation_samples}
\end{figure}

To better understand the behavior of the proposed contextual relevance metric, we additionally analyze generated massings with different metric values. For~this qualitative analysis, we use generations from the best-performing inference configuration, as~measured by the absolute contextual relevance score reported in Table~\ref{tab:top_inference_configs}: the 8B model trained in the multimodal-context regime and evaluated with $1D \times 6 + 2D + 3D \times 8$ context. Figure~\ref{fig:generation_samples} shows examples from this configuration sorted from low to high contextual relevance. The~generated building is shown in red, while surrounding buildings are shown in gray. We interpret these qualitative examples together with the feature-attribution analysis of the learned metric shown in Figure~\ref{fig:shap}.

Samples with low contextual relevance often contain strong geometric artifacts. In~the leftmost column, many generated massings have unrealistic proportions, sharp angles, or~irregular forms that differ substantially from the surrounding built fabric. Visually, this suggests that the proposed metric can act as an anomaly indicator for generated geometry. The~SHAP analysis provides a more precise explanation of this effect: density-related features, especially coverage and FAR, make a strong contribution to the learned metric. Geometric artifacts frequently disturb density-related indicators: very large, extremely thin, or~disproportionate buildings can produce abnormal ratios between built volume, footprint area, and~site area. As~a result, such samples often receive low contextual relevance~scores.

\begin{table}[H]
    \caption{Top-3 inference configurations by contextual relevance score for each model size and training strategy. Results are grouped by training regime and model size. The~table shows that the best overall configuration is the 8B model trained with multimodal context and evaluated with six 1D neighboring massing JSONs, the~2D top-down view, and~eight 3D context~views. Bold formatting highlights the best configurations for unimodal and multimodal contexts.}
    \label{tab:top_inference_configs}
    \centering
\begin{tabularx}{\textwidth}{wl{3.5cm} wl{2.5cm} X wl{2cm}}
    \toprule
    \textbf{Train Type} & \textbf{Model Size} & \textbf{Inference Type} & \textbf{CR} \\
    \midrule
    \multirow{9}{*}{Unimodal context} 
    & \multirow{3}{*}{2B} 
    & $3\mathrm{D} \times 8$ & 0.360 \\
    & & $3\mathrm{D} \times 4$ & 0.357 \\
    & & $1\mathrm{D} \times 2$ & 0.356 \\
    \cline{2-4}
    & \multirow{3}{*}{4B} 
    & $3\mathrm{D} \times 8$ & 0.407 \\
    & & $1\mathrm{D} \times 10$ & 0.400 \\
    & & $2\mathrm{D} \times 1$ & 0.399 \\
    \cline{2-4}
    & \multirow{3}{*}{8B} 
    & $3\mathrm{D} \times 4$ & \textbf{0.419} \\
    & & $3\mathrm{D} \times 8$ & 0.416 \\
    & & $3\mathrm{D} \times 6$ & 0.413 \\
    \midrule
    \multirow{9}{*}{Multimodal context} 
    & \multirow{3}{*}{2B} 
    & $1\mathrm{D} \times 6 + 3\mathrm{D} \times 8$ & 0.383 \\
    & & $1\mathrm{D} \times 10 + 2\mathrm{D} + 3\mathrm{D} \times 6$ & 0.379 \\
    & & $1\mathrm{D} \times 4 + 2\mathrm{D} + 3\mathrm{D} \times 8$ & 0.375 \\
    \cline{2-4}
    & \multirow{3}{*}{4B} 
    & $1\mathrm{D} \times 10 + 3\mathrm{D} \times 6$ & 0.429 \\
    & & $1\mathrm{D} \times 8 + 3\mathrm{D} \times 2$ & 0.428 \\
    & & $1\mathrm{D} \times 6 + 3\mathrm{D} \times 4$ & 0.427 \\
    \cline{2-4}
    & \multirow{3}{*}{8B} 
    & $1\mathrm{D} \times 6 + 2\mathrm{D} + 3\mathrm{D} \times 8$ & \textbf{0.435} \\
    & & $1\mathrm{D} \times 10 + 2\mathrm{D} + 3\mathrm{D} \times 8$ & 0.429 \\
    & & $1\mathrm{D} \times 4 + 3\mathrm{D} \times 2$ & 0.428 \\
    \bottomrule
\end{tabularx}
  \end{table}

However, the~metric should not be interpreted as a complete detector of geometric quality. Its sensitivity to artifacts is indirect and depends on whether the artifact affects features that are important for contextual matching. For~example, sample 2376 is visually plausible but receives a low score, likely because its small absolute scale produces an atypical density relationship relative to the surrounding context. Conversely, sample 4126 contains visually noticeable narrow elements, but~its overall footprint area and density remain more compatible with nearby buildings, resulting in a higher score than the visually cleaner sample. This indicates that the metric can help identify anomalous generations, but~its anomaly-detection accuracy is limited: some geometric defects may not strongly affect contextual features, while contextually unusual but geometrically valid buildings can still receive low~scores.

The examples also demonstrate that the metric is sensitive to the relationship between a generated massing and its specific context, rather than only to intrinsic shape quality. Similar buildings can receive different contextual relevance values depending on their surroundings. For~instance, samples 455 and 3465 are both close to rectangular massings, but~sample 3465 receives a higher contextual relevance score because it is surrounded by buildings with more uniform height and simpler forms. In~contrast, the~neighboring buildings around sample 455 have a more stepped morphology, while the generated massing itself remains comparatively simple, making it less contextually aligned. This interpretation is consistent with the feature-attribution analysis: height-related features and elevation diversity have a noticeable, although~not dominant, influence on the metric. Overall, these examples support the intended role of contextual relevance as a proxy for morphological correspondence between generated massing and its surrounding urban context, rather than as a purely visual or geometric-quality~score.

\section{Discussion}
\unskip

\subsection{Limitations}

This study has several assumptions and limitations. Firstly, the~dataset is restricted to one city, so both the generated results and the learned contextual relevance metric may reflect Melbourne-specific urban patterns. The~particularly high importance of orientation in a benchmark constructed using randomly sampled mismatched contexts may also be partly amplified by the sampling procedure, because~directional mismatches may be easier to identify than subtler differences in height, density, or~footprint shape. Orientation can therefore be regarded as an established morphological signal, while its relative dominance in the present benchmark should not be assumed to generalize unchanged beyond Melbourne. The~orientation-feature ablation in Appendix~\ref{app:metric_ablation} confirms this limitation: removing orientation weakens the learned metric but does not reverse the main conclusions about multimodal training and inference. Future work should evaluate the method on broader multi-city datasets, including cities with irregular street networks and substantially different urban morphologies, to~test how strongly the learned metric and generation results depend on the individual city~morphology.

\textcolor{black}{Secondly, the~metric itself is also only an approximation of morphological contextual compatibility: it relies on the available massing-level features, and~its agreement with independent expert assessment of architectural style has not yet been established. The~negative examples used for metric selection are uniformly sampled foreign contexts. Because~Melbourne and most cities in general contain neighborhoods with similar orientation, density, height, and~footprint morphology, such negatives are not always trivially separable; nevertheless, the~robustness of the CR metric to deliberately hard negatives matched on these properties has not yet been demonstrated and would require a stricter future benchmark.}

Thirdly, we only use the Qwen3-VL architecture in our experiments, so our findings allow us to make hypotheses about the behavior of VLM models, but~they require further verification on a larger number of modern architectures. Finally, we only explore the contribution of modalities in the task of contextual massing generation, so further research is required to generalize our assumptions to other generative design~tasks.

These limitations define the intended scope of the study. However, despite them, the~results of the experiment allow us to draw conclusions and make assumptions that are useful for answering our research~questions.

\subsection{Implications}

Our observations allow us to propose design principles for VLM-based systems for generative design tasks that require the analysis of information presented in vector modalities, such as CAD parts or BIM models. VLMs enable natural language-based design control and image analysis, and~our findings suggest adaptations for handling parametric geometry. These systems can be used as engineering software assistants, providing a natural language interface for interacting with software commands that can analyze and edit geometry. In~addition, with~a large number of high-quality design projects, VLM can be used not only as an assistant but also as a tool for designing from scratch, and~our principles can improve the quality of such~systems.

\section{Conclusions}

This paper studied contextual massing generation as a multimodal structured-output task for vision-language models. We focused on how VLMs use surrounding urban context represented through different modalities: textual JSON descriptions of neighboring massings, a~top-down map view, and~multi-view 3D images. To~support this analysis, we assembled an experimental dataset of Melbourne building massings, implemented a context-aware VLM fine-tuning and inference pipeline, and~introduced a contextual relevance metric for estimating morphological correspondence between generated massings and surrounding urban~form.

The experiments show that model capacity strongly affects contextual relevance, while the way context is represented and combined also plays a central role. Multimodal training improves the use of individual context modalities, and~multimodal inference gives the strongest results under the proposed contextual relevance metric and tested configurations. This suggests that, for~VLMs, visual and vector representations of urban context are not simply interchangeable encodings of the same information. Instead, they provide complementary signals that help the model interpret spatial context and generate more contextually aligned massing~geometry.

These findings may inform future work beyond the specific dataset and metric used in this study. They suggest that VLM-based generative design systems can benefit from combining editable vector representations with visual context, especially when working with architectural and urban geometry. Future work should address the limitations of the current setup by extending the dataset to more diverse urban environments, considering a broader set of generation tasks and contextual metrics, validating these metrics through human evaluation, and~comparing different VLM architectures to understand how architectural choices affect the use of multimodal spatial~context.

\vspace{6pt}

\authorcontributions{E.M. conceived the experimental idea, collected and prepared the data, implemented and ran the training and inference pipelines, and~drafted the manuscript. A.V. (Alexandra Vabnits) developed the contextual relevance metric, implemented the validation code, and~drafted the manuscript. V.V. contributed to the evaluation methodology, assisted with metric development and validation code implementation, and~conducted the literature review. A.A. contributed to the research proposal, analyzed the results, drafted the manuscript, and~reviewed and edited the manuscript. V.K. contributed to the experimental methodology, assisted with training and inference runs, and~reviewed and edited the manuscript. A.V. (Anastasia Volkova) assisted with data collection and reviewed and edited the manuscript. A.G. contributed to the task formulation and reviewed and edited the manuscript. A.K. administered the project. I.O. provided scientific supervision. All authors have read and agreed to the published version of the manuscript.}

\funding{This work was supported by the Ministry of Economic Development of the Russian Federation (Agreement No. 139-10-2025-034 dated June 19, 2025, unique identifier IGK 000000C313925P4D0002).}

\dataavailability{The original contributions presented in this study are included in the Supplementary Materials. Further inquiries can be directed to the corresponding author.} 

\supplementary{The training dataset, model weights, and inference results are available at: \url{https://zenodo.org/records/21344986}. The code is available at: \url{https://github.com/FusionBrainLab/CoMa}.}

\acknowledgments{The authors acknowledge the contribution of the Institute of Design and Urban Studies at ITMO University to the review of existing methods and the development of evaluation metrics used in this study.}

\conflictsofinterest{The authors declare no conflicts of~interest.}

\appendixtitles{yes}
\appendixstart
\appendix

\section{Analysis of Spatial Context~Overlap}
\label{app:context_overlap}
\unskip

\subsection{Addressing Train--Test Context~Overlap}

The dataset was constructed from a single city and divided into training and test sets using a random split. This protocol was selected to evaluate in-distribution generalization while preserving similar distributions of building morphology, development-site geometry, and~urban typology in both~subsets.

We acknowledge that a random split does not ensure spatial independence. Training and test sites may belong to the same neighborhood or even be adjacent. Consequently, their surrounding urban contexts can partially overlap and are also likely to share similar morphological~characteristics.

Morphological similarity alone is expected in an in-distribution evaluation: a model can only be expected to generalize reliably to examples represented by, or~sufficiently close to, its training distribution. Exact reuse of the same context buildings is a different concern. If~the model repeatedly observes the same buildings in both training and test contexts, the~reported performance could partly reflect context memorization rather than generalization to unseen contextual configurations. We therefore conducted an additional ablation to estimate the effect of exact context~overlap.

This analysis addresses overlap within the random split. It does not constitute a spatially independent evaluation and should not be interpreted as evidence of generalization to unseen neighborhoods or~cities.

We restrict the quantitative overlap analysis to the 1D context. In~this representation, every context element is associated with an explicit dataset identifier, making it possible to reconstruct exactly which buildings were presented to the~model.

The same analysis is less direct for the image modalities. The~2D representation is rasterized from map data and does not retain an explicit correspondence between image regions and dataset building identifiers. For~3D renderings, the~buildings visible in a particular image additionally depend on camera orientation, field of view, building height, and~occlusion. Determining exact overlap for these modalities would therefore require a separate visibility-aware reconstruction. Accordingly, all results below concern inference configurations with a non-zero 1D~context.

Candidate context buildings were initially selected within a radius of 100\,m around the target development site. However, the~model did not receive every building within this radius. During~both training and inference, only a subset of candidate buildings was~sampled.

\subsection{Context Overlap Ablation and~Results}

Let
\begin{equation}
    C_s = \{b_1,\ldots,b_m\}
\end{equation}
denote the sampled 1D context of test sample $s$. For~every unordered pair of context buildings $(b_i,b_j)$, we count the number of retained training samples in which both buildings were simultaneously present:
\begin{equation}
    n_{ij}
    =
    \left|
    \left\{
    t :
    b_i \in C_t^{\mathrm{train}}
    \land
    b_j \in C_t^{\mathrm{train}}
    \right\}
    \right|.
\end{equation}

This definition differs from checking whether the individual buildings appeared anywhere in training. A~pair receives a positive count only when both buildings were jointly presented as part of the same training~context.

To account for repeated exposure while preventing highly frequent pairs from dominating the score, the~occurrence counts are transformed using a saturating function:
\begin{equation}
    e_{ij}
    =
    1 - \exp\left(-\frac{n_{ij}}{\tau}\right),
\end{equation}
where $\tau$ is the median positive pair-occurrence count in the corresponding training~dataset.

The co-exposure score of test sample $s$ is defined as
\begin{equation}
    \operatorname{CoExposure}(s)
    =
    \frac{1}{\binom{m}{2}}
    \sum_{i<j} e_{ij}.
\end{equation}

The score equals zero when none of the context pairs were observed jointly during training and approaches one when most pairs were observed repeatedly. Samples containing fewer than two valid 1D context buildings are assigned a score of~zero.

\begin{figure}[H]

    \includegraphics[width=\linewidth]{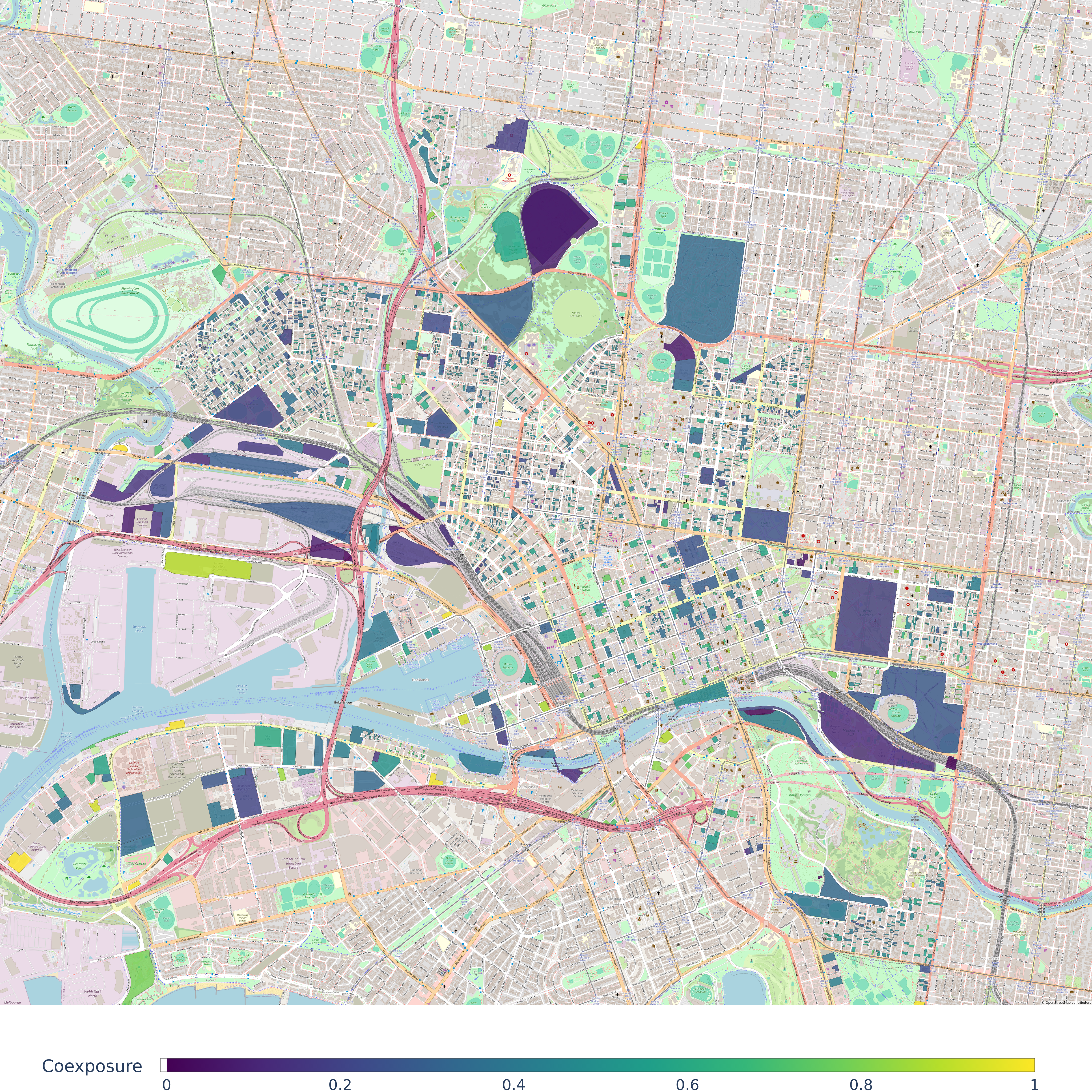}
    \caption{Spatial distribution of pairwise context co-exposure across test sites. The~color of each development site represents the average repeated exposure of pairs of its sampled 1D context buildings in the corresponding training~dataset.}
    \label{fig:context_coexposure_map}
\end{figure}

Figure~\ref{fig:context_coexposure_map} visualizes the spatial distribution of the co-exposure score. The~variation across test sites demonstrates that context overlap is not uniform: despite the spatial intermixing produced by the random split, some test contexts are substantially more exposed during training than~others.

To determine whether context overlap systematically affects model quality, we ranked samples separately within every inference configuration according to their co-exposure~score.

For each configuration, we formed two~subsets:
\begin{itemize}
    \item Lower-overlap subset: the bottom 20\% of samples according to co-exposure;
    \item Upper-overlap subset: the top 20\% of samples according to co-exposure.
\end{itemize}

The subsets were formed independently for every model size and context configuration. This prevents differences in model capacity or in the number of supplied context elements from affecting the~ranking.

We recomputed the evaluation metric on each subset and compared it with the result obtained on the complete inference dataset:
\begin{equation}
    \Delta M_{\mathrm{subset}}
    =
    M_{\mathrm{subset}}
    -
    M_{\mathrm{full}}.
\end{equation}

A systematic memorization effect would be expected to produce positive deviations for the upper-overlap subsets and negative deviations for the lower-overlap~subsets.

\begin{figure}[H]

    \includegraphics[width=\linewidth]{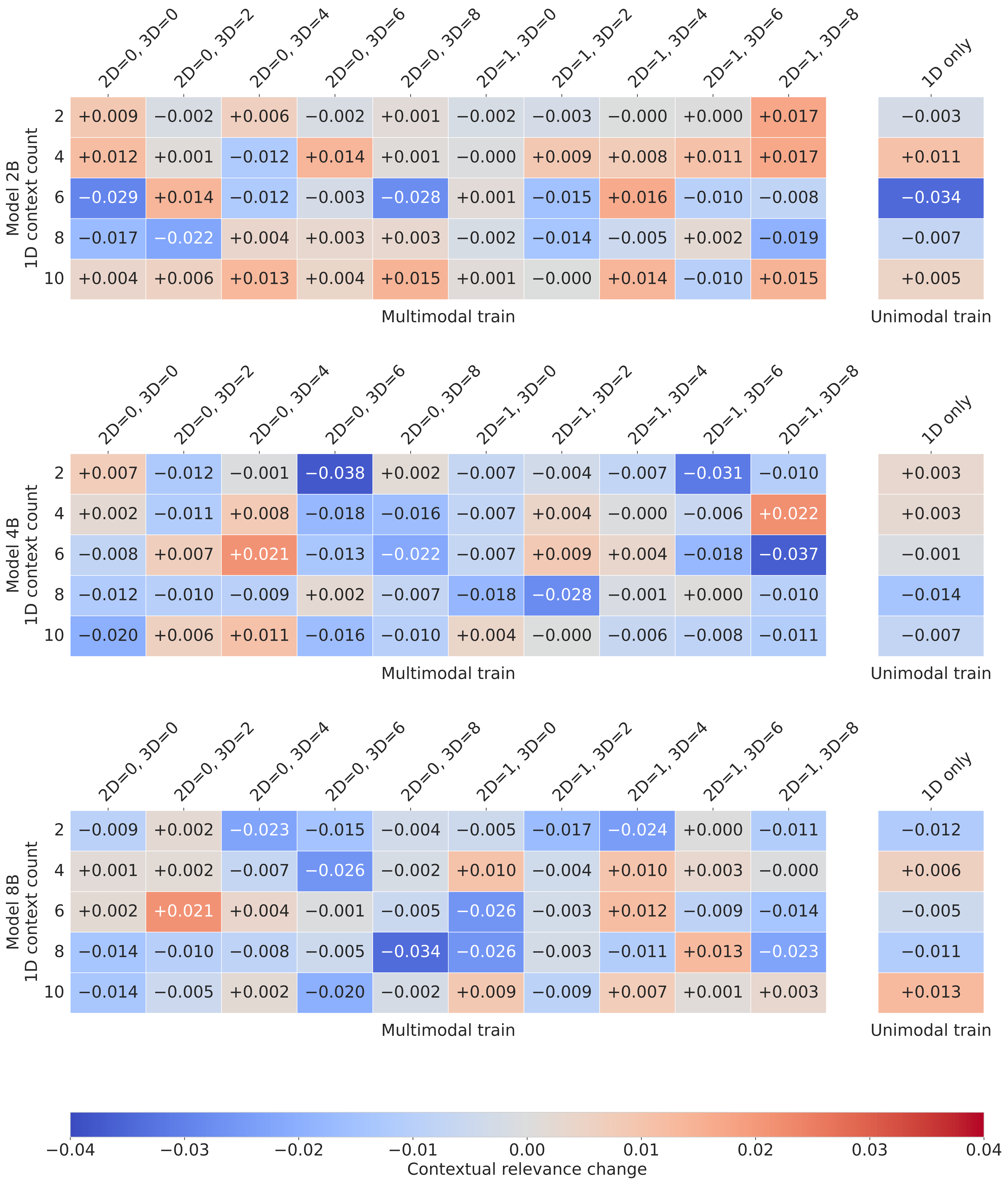}
    \caption{Change in contextual relevance on the bottom 20\% of test samples according to pairwise context co-exposure. Each value is computed relative to the corresponding complete inference~dataset.}
    \label{fig:lower_overlap_delta}
\end{figure}

\begin{figure}[H]

    \includegraphics[width=\linewidth]{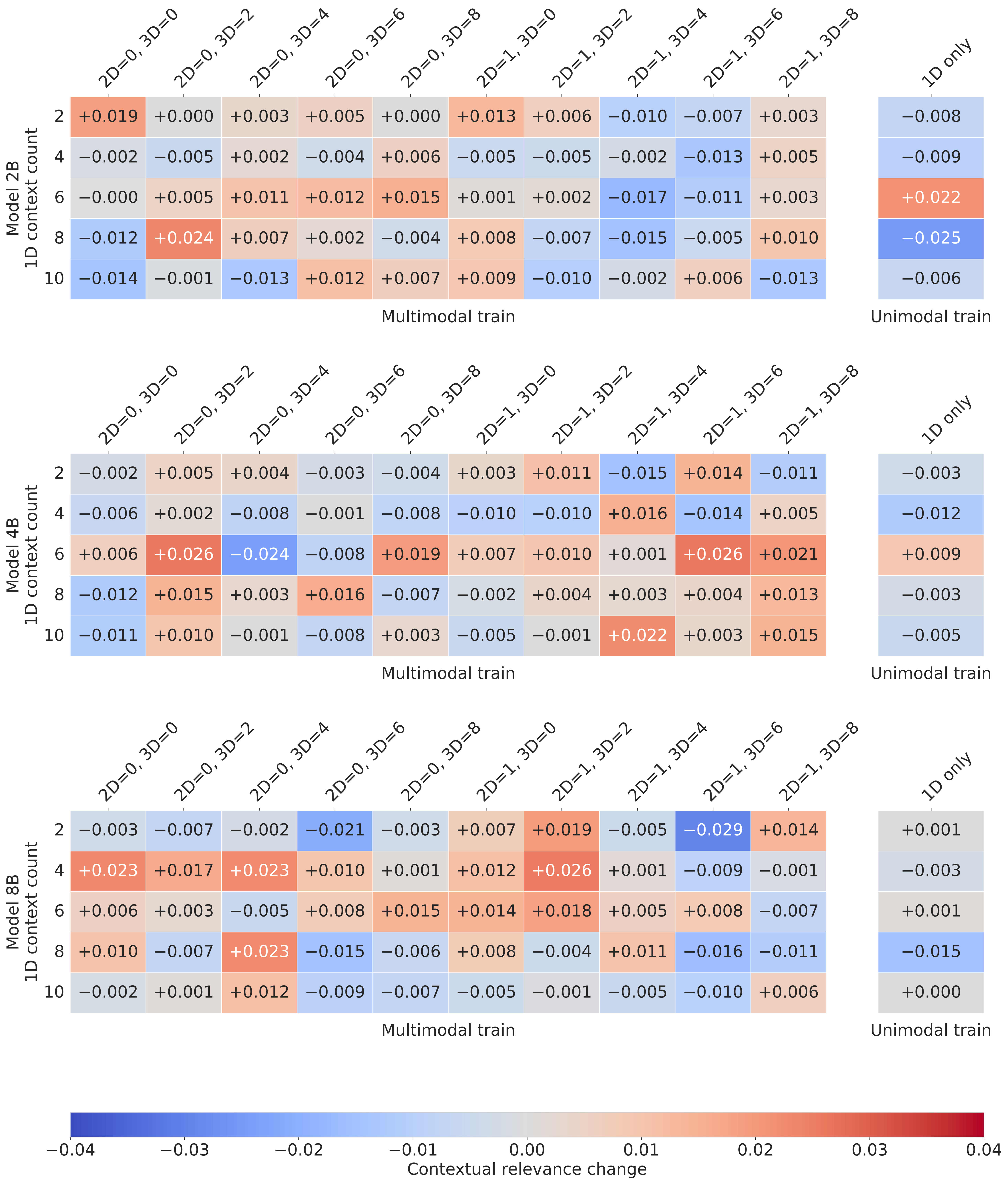}
    \caption{Change in contextual relevance on the top 20\% of test samples according to pairwise context co-exposure. Each value is computed relative to the corresponding complete inference~dataset.}
    \label{fig:upper_overlap_delta}
\end{figure}

Figures~\ref{fig:lower_overlap_delta} and~\ref{fig:upper_overlap_delta} report the metric deviations for the lower- and upper-overlap subsets, respectively, across model sizes and context~configurations.

The results reveal a weak but consistent association between train--test context overlap and model performance. On~average, contextual relevance increases for the upper co-exposure subset and decreases for the lower subset. This indicates that samples whose contexts were more frequently observed during training tend to be evaluated more favorably. However, the~trend is not uniform across submissions: both heatmaps contain numerous configurations whose metric changes are negligible or opposite in sign. Therefore, context overlap affects the reported performance, but~it explains only a limited fraction of the variation and does not appear to be the dominant~factor.

This conclusion is limited to exact overlap in the sampled 1D context. It does not rule out effects caused by morphological similarity, overlap in image-based context, or~broader neighborhood-level correlations. A~geographically disjoint split would remain a stronger test of spatial and cross-neighborhood~generalization.

\section{Orientation-Feature Ablation of the Contextual Relevance~Metric}
\label{app:metric_ablation}

The feature analysis of the contextual relevance (CR) metric shows that orientation-related features are highly predictive; for example, $\mathrm{Dir}_{\mathrm{near}}$ reaches an ROC-AUC of 0.861. To~verify that the conclusions about multimodal training and inference are not driven only by orientation alignment, we retrained the learned CatBoost CR metric after removing all features that explicitly encode global footprint orientation: $\mathrm{Dir}_{\mathrm{near}}$, $\mathrm{Dir}_{\mathrm{mean}}$, $\mathrm{Fr}_{\mathrm{orient}}$, $\mathrm{KDE}_{\mathrm{orient}}$, and~$\mathrm{Fr}_{\mathrm{multi}}$. The~last feature was removed because its ``classic'' Frechet subset includes $\sin(2\theta)$ and $\cos(2\theta)$ terms. We retained $\mathrm{AgD}_{\mathrm{near}}$, because~it measures rotation-invariant polygonal angle diversity rather than global~orientation.

Removing orientation substantially weakens the learned metric, confirming that orientation is an important component of contextual relevance in this dataset (Table~\ref{tab:metric_orientation_ablation}). Nevertheless, the~ablated metric remains well above chance, showing that scale, density, height, compactness, elongation, concavity ratio, extrusion diversity, elevation diversity, setback, stepback, and~angle-diversity features still encode contextual~compatibility.

\begin{table}[H]
    \caption{Effect of removing orientation-related features from the learned contextual relevance metric. Values report grouped cross-validation on the metric-validation~benchmark.}
    \label{tab:metric_orientation_ablation}
    \centering
    \begin{tabularx}{\linewidth}{Xcc}
        \toprule
        \textbf{Learned CR metric} & \textbf{ROC-AUC} & \textbf{F1} \\
        \midrule
        All 17 features & $0.915 \pm 0.003$ & $0.848 \pm 0.006$ \\
        Without orientation features & $0.762 \pm 0.005$ & $0.733 \pm 0.003$ \\
        \bottomrule
    \end{tabularx}
\end{table}

\begin{figure}[H]

    \includegraphics[width=\textwidth]{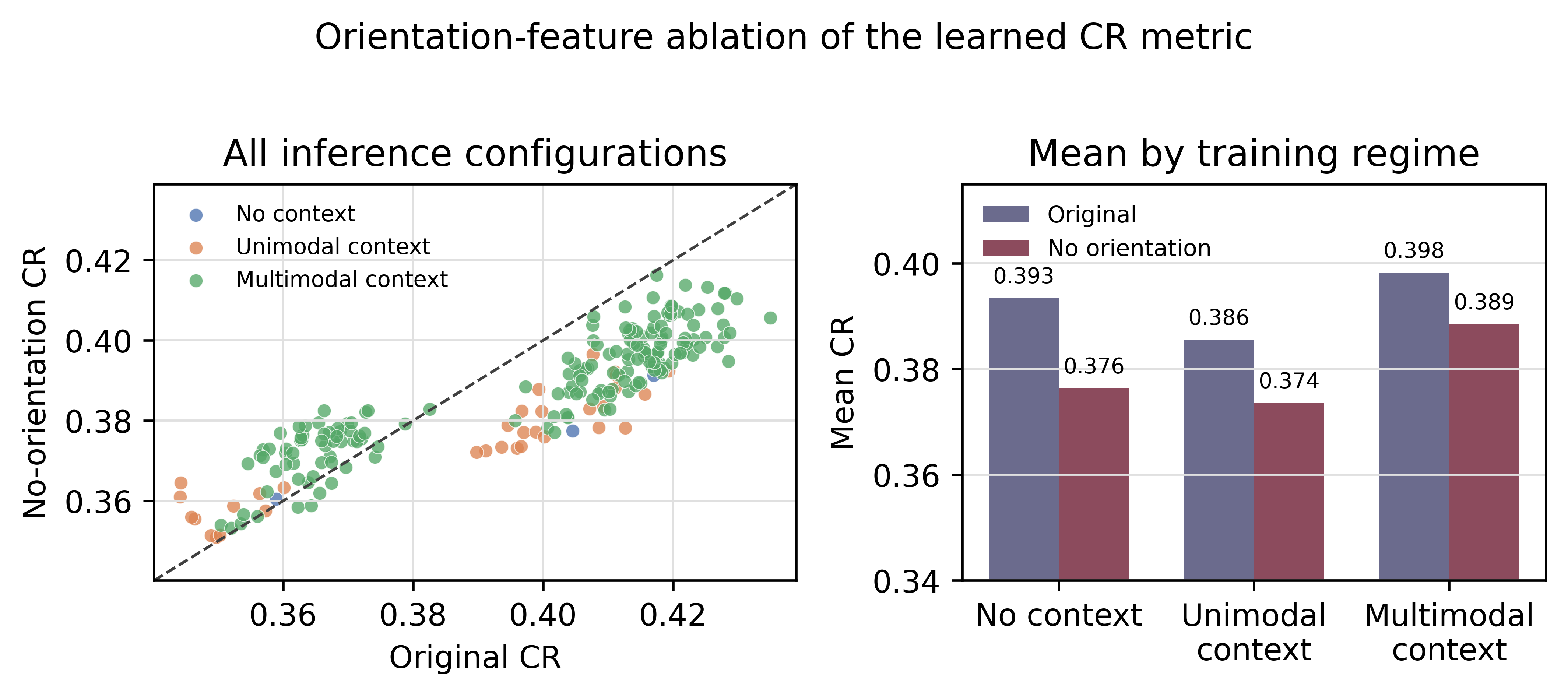}
    \caption{Comparison between the original CR metric and the no-orientation ablation. Left: all 216 inference configurations, with~the dashed line denoting equal scores under both metrics. Right: mean CR score by training regime. The~ablated metric lowers the absolute score scale, but~the ordering of the main training regimes remains~consistent.}
    \label{fig:ablation_summary}
\end{figure}

We then rescored all 216 generation-evaluation configurations with the no-orientation metric. As~expected, scores become lower and more compressed: the mean score changes from 0.396 to 0.386, and~the best score changes from 0.435 to 0.416. Figure~\ref{fig:ablation_summary} summarizes this comparison. In~the left panel, each point corresponds to one inference configuration. The~dashed diagonal marks equal scores under the original and ablated metrics; most points fall below it, showing the expected score decrease after orientation removal. At~the same time, the~cloud remains structured rather than random: configurations that score high under the original metric usually remain high under the ablated metric. This indicates that the ablation changes the score scale more than it changes the qualitative ordering of configurations. The~right panel shows the same conclusion after averaging by training regime. Multimodal-context training remains strongest under the ablated metric (0.389), followed by no-context training (0.376) and unimodal-context training (0.374).

\begin{figure}[H]

    \includegraphics[width=\textwidth]{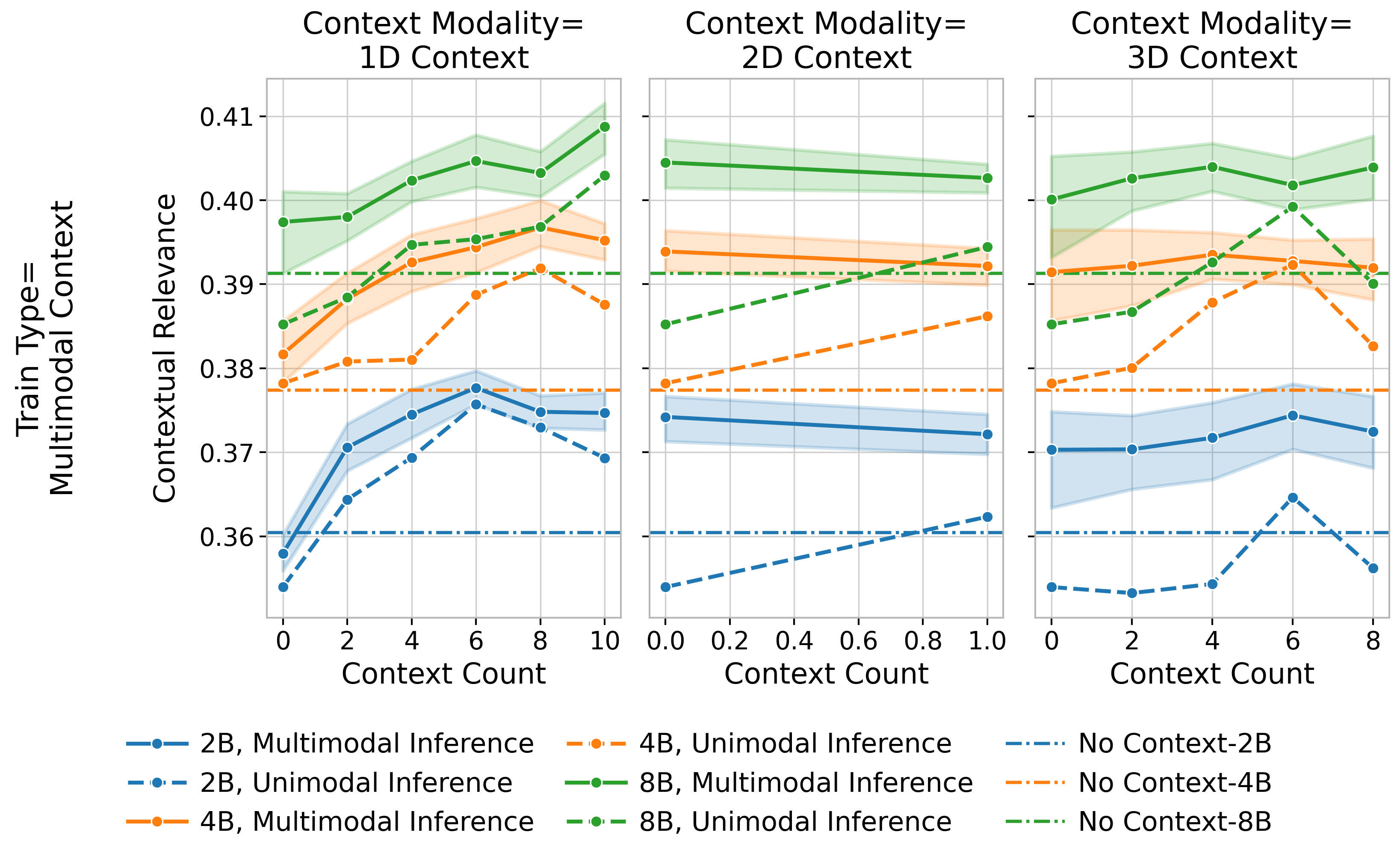}
    \caption{No-orientation ablation: comparison of single-modality and multimodal inference for multimodally trained models. The~plot is the ablated counterpart of Figure~\ref{fig:multimodal_line_grid}. Combined context modalities remain beneficial under the orientation-free~evaluator.}
    \label{fig:ablation_multimodal_line_grid}
\end{figure}

Figure~\ref{fig:ablation_multimodal_line_grid} focuses on multimodally trained models and directly tests the inference-side conclusion. It compares single-modality inference with configurations, where the same modality is combined with other context sources. The~solid lines, corresponding to multimodal inference, generally remain above the short-dashed single-modality lines. This supports the same conclusion as the main analysis: vector geometry, top-down imagery, and~3D views provide partly complementary information. The~effect is modest in absolute value because the ablated metric has a compressed scale, but~it is consistent across the 4B and 8B models. For~the full multimodal inference setting ($1D \times 10 + 2D + 3D \times 8$), the~gain over the no-context baseline under the ablated metric is $+0.014$, $+0.021$, and~$+0.019$ for the 2B, 4B, and~8B models, respectively. These gains are not smaller than under the original metric ($+0.010$, $+0.010$, and~$+0.013$).

Overall, the~ablation confirms that orientation is an important component of the learned CR metric, but~it does not reverse the substantive conclusions. Removing orientation lowers the absolute discriminative strength of the evaluator and compresses score differences, yet multimodal training remains advantageous, multimodal inference still benefits from combining context representations, and~larger models still obtain higher contextual~relevance.

\vspace{-9pt} 

\reftitle{References}

\PublishersNote{}

\end{document}